# Advanced Intelligent Systems for Surgical Robotics

*Mai Thanh Thai [#], Phuoc Thien Phan[#], Shing Wong, Nigel H. Lovell, and Thanh Nho Do**

*[#] Equal Contributions*

** Corresponding Author(s)*

Mr. Mai Thanh Thai, Mr. Phuoc Thien Phan, Prof. Nigel H. Lovell, and Dr. Thanh Nho Do

Graduate School of Biomedical Engineering, Faculty of Engineering, UNSW, Kensington Campus, Sydney, NSW 2052, Australia.

*E-mail*: tn.do@unsw.edu.au

Dr. Shing Wong

Prince of Wales Hospital, Randwick, Sydney, NSW 2031, Australia

Faculty of Medicine, UNSW, Kensington Campus, Sydney, NSW 2052, Australia

**Abstract**

Surgical robots have had clinical use since the mid 1990s. Robot-assisted surgeries offer many benefits over the conventional approach including lower risk of infection and blood loss, shorter recovery, and an overall safer procedure for patients. The past few decades have shown many emerging surgical robotic platforms that can work in complex and confined channels of the internal human organs and improve the cognitive and physical skills of the surgeons during the operation. Advanced technologies for sensing, actuation, and intelligent control have enabled multiple surgical devices to simultaneously operate within the human body at low cost and with more efficiency. Despite advances, current surgical intervention systems are not able to execute autonomous tasks and make cognitive decisions that are analogous to that of humans. This paper will overview a historical development of surgery from conventional open to robotic-assisted approaches with discussion on the capabilities of advanced intelligent systems and devices that are currently implemented in existing surgical robotic systems. It will also revisit available autonomous surgical platforms with comments on the essential technologies, existing challenges, and suggestions for the future development of intelligent robotic-assisted surgical systems towards the achievement of fully autonomous operation.

## 1. Introduction

Surgery (Ancient Greek: χειρουργία) is a branch of medicine that can deal with the physical manipulation of a bodily structure to diagnose, prevent, or cure an ailment [1-2]. The first evidence of a surgical procedure occurred as early as 3000 B.C. The practice of surgery, thereafter, was widespread throughout South America, Europe, and Asia although the initial purpose of these surgeries is unknown [1-2]. Open surgeries have been used to treat diseases for many years. With better surgical techniques, better analgesia and better infection control during operations, open surgeries with a large skin incision are still associated with "dreadful, painful, and dangerous feelings" [3].

Recently, modern surgeries with miniature tools have demonstrated many benefits over conventional open incisions with shorter recovery time, low complication rates, and better cosmesis [4]. These technologies play an important role for global healthcare and have been considered as indispensable methods for a wide range of treatments, to mitigate pain that patients may suffer from in malignant diseases involving gastrointestinal tracts, lung, liver, bladder, and prostate. They can treat other medical conditions such as musculoskeletal diseases, faulty body organs (kidney stones, gallbladder, or appendix) and removal of foreign bodies (shrapnel or bullets) [5]. Minimally invasive surgery (MIS) have gathered more attention in surgical communities during the past few years as it overcomes many drawbacks of conventional “open” surgical procedures. MIS offers many benefits such as a reduction in patient trauma, less blood loss, shorter healing time,improvement in diagnostic outcomes, better cosmesis, and faster recovery [6-9]. However, the visual and haptic feedback to surgeons in MIS is impaired. One of the MIS techniques is laparoscopic surgery. This method approaches the internal organs via a tiny camera integrated at the laparoscope and miniature instruments inserted via ports in small skin keyhole incisions [10-11]. It is normally performed under general anesthesia by a surgeon or gynaecologist (women's health specialist) [12]. Despite advances, laparoscopic surgery still has several drawbacks compared with conventional open surgery. This includes impaired haptic and visual feeback. The laparoscopic surgeon dissects the tissue by navigating surgical instruments from a distance with no force feedback and the visual feedback from a camera does not provide target depth information because the picture is two-dimensional. In addition, laparoscopic surgery requires extra training to manipulate surgical tools and there is a learning curve to attain technical expertise [13]. Recent development in actuation technology with miniature cable or magnetic field has advanced the use of surgical tools to mimic the movements and dexterity of the human hand and wrist with a high number of degrees of freedom (DOFs). Surgical tools can be scaled down to few milimeters in size that allows the surgeons to precisely manipulate these tools within complex and narrow paths of the human organs and through small skin incisions [14]. Advances in visual technology also offer high-resolution 3-dimensional real-time videos from the integrated camera to provide more precise views of target tissues resulting in improved outcomes [15].

The era of modern surgical systems has initiated revolutions to overcome limitations in conventional "open" surgical procedures resulting in reduction of side effects, enhanced surgical precision, small skin incisions, and faster patient recovery. One of many applications of MIS is Natural Orifice Transluminal Endoscopic Surgery (NOTES). NOTES is a new paradigm that makes the use of human natural orifices to access the abdominal cavity for surgery without leaving any visible scars [16-17]. Access for NOTES can be achieved via transvaginal, transoral, and transanal avenues. With advances in actuation and sensing technologies such as the development of miniature size and high force instruments, more advanced NOTES operations would be possible [18-22]. NOTES procedures significantly reduce pain, leave no visible scars, and result in faster recovery. NOTES was first published in a report in 2004 by Kalloo et al. from the Johns Hopkins Hospital [22]. High maneuverability with improvements in articulation and triangulation are important features for successful NOTES surgery. Besides its advantages, NOTES is technically demanding over conventional open and laparoscopic surgeries. With spatial constraints within very tight, long, and narrow channels, current available technology of mechanical tools for NOTES is inadequate to enhance tracking performances and provide haptic feedback to the surgeons [23-27]. The surgical tools need to have a high flexibility within a compact space and sufficient dexterity to carry out surgical tasks through the human natural orifices. In addition, clear vision signals from the camera and fidelity of haptic feedback to the surgeons are needed to assist safe surgery [28-32].

## 2. Robot-Assisted Surgery

Advances in modern technologies in recent years have revolutionized the field of surgery. Cutting-edge robot-assisted minimally invasive procedures are now increasingly adopted into clinical practice to substitute for conventional open surgery [33-36]. The trend in general surgery has moved towards less invasive procedures with an increase in the use of laparoscopic and robotic surgery. Robot-assisted surgery is a new type of surgical procedures that is performed via the use of robotic systems. This type of surgery was developed to overcome major limitations of existing MIS procedures in order to enhance the capabilities of surgeons [37]. A surgical robot can be defined as computer controlled/ assisted dexterous tools with advanced sensing and display technologies that can be motion programmed to carry out surgical tasks [38-39]. The general intention of the use of a surgical robot is to assist the surgeon during the surgery. The most common control of surgical robot is a repetitive motion that maps the surgeon hand’s movements to the motion of the surgical tools [40].

The first robot-assisted surgery was performed in 1985 to conduct a neurosurgical procedure which required delicate precision while the first robot-assisted laparoscopic cholecystectomy was carried out in 1987 [41]. In robot-assisted MIS, the surgeon uses a tele-manipulation system to control surgical arms. Surgery is performed from a remote site via a master console which manipulates the motion at the distal end via a motor housing that is directly connected to the arms. The robotic arms follow exactly the movements of the surgeon to perform the actual surgery within the human body [28, 42-43]. Another type of robot-assisted surgery is a computer-controlled robotic system [44-46]. In this system, the surgeon uses a computer program to control the robotic arms and its end-effectors with an advantage of providing more assistance via machine learning approaches, virtual reality, haptics, and artificial intelligent (AI). For the computerized surgical method, surgeons can remotely control the system from a distance, leading to the possibility for remote surgery from anywhere [47-48].

Despite early success in this field, robotic surgery did not join the surgical mainstream until the early 2000s when the da Vinci system (Intuitive Surgical, CA, USA) was first approved by the USA Food and Drug Administration (FDA). This system was also set as a bar for robot-assisted surgery and it is currently one of the most common platforms of robotic surgery in the world [49]. Subsequently, Mimic and Intuitive Surgical Inc released the da Vinci Skills Simulator that is specifically designed to give surgeons the opportunity to improve their proficiency before carrying out real operations [50]. Although other surgical systems have been developed to provide additional surgical options for the patients such EndoMaster, Medrobotics, Microbot Medical, and Titan Medical, the Da Vinci system is still the most popular with over 875,000 da Vinci procedures performed in 2013, up from 523,000 [33]. These emerging systems will offer a variety of choices, versatility and functions in the competitive surgical sector market.

Typically, a surgical system consists of a remote console where a surgeon commands the remote robotic arms via a master controller in a non-sterile section of the operating room. The robotic arms are slave wrist-mimicking instruments placed in a patient cart and provide visual displays. In most existing systems, surgeons perform surgical procedures via a direct mapping that transform their hands' movements at the master console to the bending and rotating motion of surgical instruments. Depend on the type of procedure, surgical robotic system can be performed via a laparoscopic or NOTES approach. Advances in visual displays also enable 3D images from a binocular camera system to be displayed, allowing the surgeon to determine the relative position between the instrument tips and organ tissue more precisely, and hence significantly enhancing the accurate control of the operating tool tip [51-52].

The increasing use of robots in surgery has resulted in shorter operating times, resulting in better outcomes for both the patients and surgeons. Surgeons have seen many benefits in ergonomics with a console that allows them to sit while operating compared with standing with shoulders and arms held in unnatural positions for a long period of time with laparoscopic surgery [53]. In addition, robotic approaches may result in economic benefits from potentially lower costs compared with laparoscopic surgery [54].

The use of master a console with joystick controls compensates for the fulcrum effect at the skin level with the the use of laparoscopic instruments. The tip of the instruments move in the same direction as the hand movements in robotic surgery, whereas they move in opposite directions in laparoscopic surgery. In this way, robotic surgery mimics open surgery more closely and may be easier to learn. The better visual display magnifies the view of the surgical target during the procedure. The advantages of robotic surgical systems include safer operations; minimization of visible skin incisions; shortened recovery time; faster and more accurate surgery, and reduced postoperative pain [55]. The robot holds the ports steady at the skin level which results in less pressure and trauma on the abdominal wall.

Recent surgical systems also integrate optical visualization in 3D to enhance surgical maneuverability and accuracy for retraction, exposure, and resection of tissue. In addition, the revolutionization of wireless technologies enhances the instant communication between patients and clinicians via tele-health of medical data leading to improved diagnosis and treatment [56]. Wide ranges of experience can be gained from the large variety of cases performed with robotic assistance. This includes surgery for prostate cancer, spinal disease, radiosurgery, orthopedic conditiond, pancreatectomy, heart surgery, bowel resection, and cardiac catheter ablation. In tele-operated surgical systems the master console is now available with real-time haptic display which can receive the force information from the distal sensor and then provide feedback to the surgeon so that accurate manipulation of the arms can be achieved when interacting with the tissue (Omega Haptic Devices [57]). The robotic instruments provide better wrist articulation compared to conventional hand-help laparoscopic instruments [58-62].

Researchers have developed a new generation of commercial laparoscopic surgical robots. They include ZEUS and Da Vinci surgical systems, RAVEN and MiroSurge robots, FreeHand and Telelap ALF-X teleoperated surgical systems, NeroArm and MrBot robots, ARTEMIS for cardiac surgery, flexible endoscopic systems such as Transport (from USGI medical), Cobra (from USGI Medical), NeoGuide (from NeoGuide Systems Inc.), ANUBISCOPE (from Karl Storz), R-Scope (from Olympus), EndoSamurai (from Olympus), DDES system (from Boston Scientific), Incisionless Operating Platform-IOP (from USGI Medical), SPIDER (from TransEnterix), ViaCath (from Hansen Medical), and MASTER robots (from Nanyang Technological University, Singapore), HVSPS system (from Munich Technological University), Single Port System (from Waseda University), HARP (from CMU), i-Snake (from Imperial College London), and IREP (from Vanderbilt University). Details of these systems are given in **Table 1**.

## 3. Advanced Intelligent Systems for Robot-Assisted Surgery

The Da Vinci system after many years of domination are now facing market competition from many international companies with new generations of surgical robots [63]. Better imaging and haptic display, better ergonomic master console, smaller instrument, and greater portability will be the main priorities in the development of new surgical robots. There is also a new trend for the automation of surgical systems when the surgeon can teleoperate the system from a remote distance under the support of advanced intelligent systems. The next generation robots are associated with faster digital communication, better decision-making abilities, enhanced visual displays and guidance, and haptic feedback. This section will discuss advance of technology-assisted current surgical robotic systems as shown in **Figure 1**.

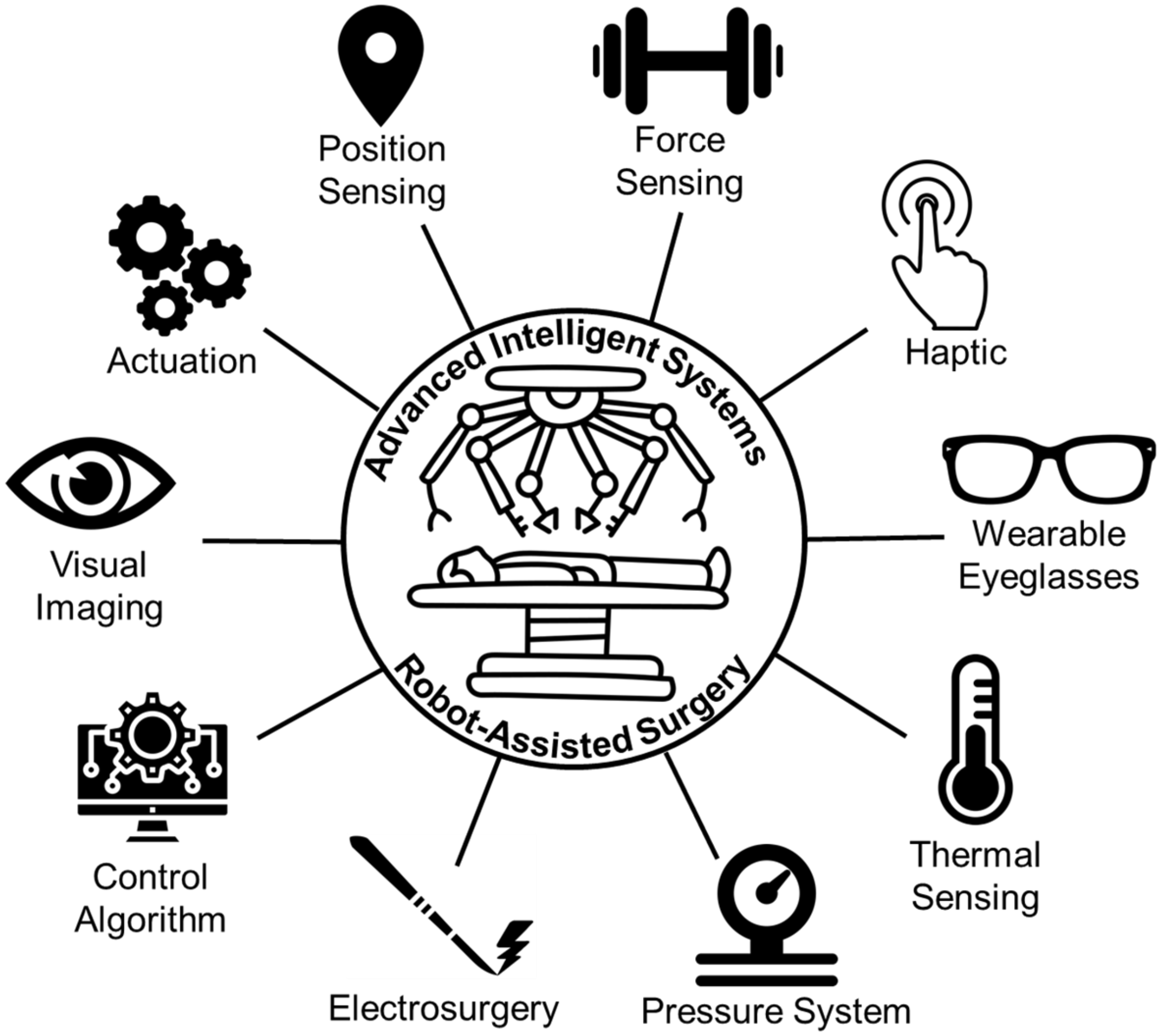

Figure 1. Advanced intelligent systems for surgical robots

### 3.1. Visual imaging for surgical robotics

Visual imaging and robotic technologies have contributed substantially to the diagnosis and treatment of many human diseases, especially in the field of surgery. Many imaging technologies such as medical ultrasound (US), computed tomography (CT), magnetic resonance imaging (MRI), and endoscopic images have been developed and successfully implemented [64]. These medical imaging devices serve as additional assistive eyes to the surgeon whne the anatomical information is not visual to the naked eye. Figure 2 illustrates three prevalent image modalities that can recognize any suspicious regions using the US, CT, and MRI technologies [65] . Despite advances, different imaging methods has their own specific merits and drawbacks and therefore surgeons normally make recommendation on which imaging method to use for the diagnosis and treatment of the patient. Advanced imaging technologies are being increasingly implemented in many clinical applications where the needs of real-time information, ease of use, and lower drawbacks of conventional radiological imaging are highly desirable [65].

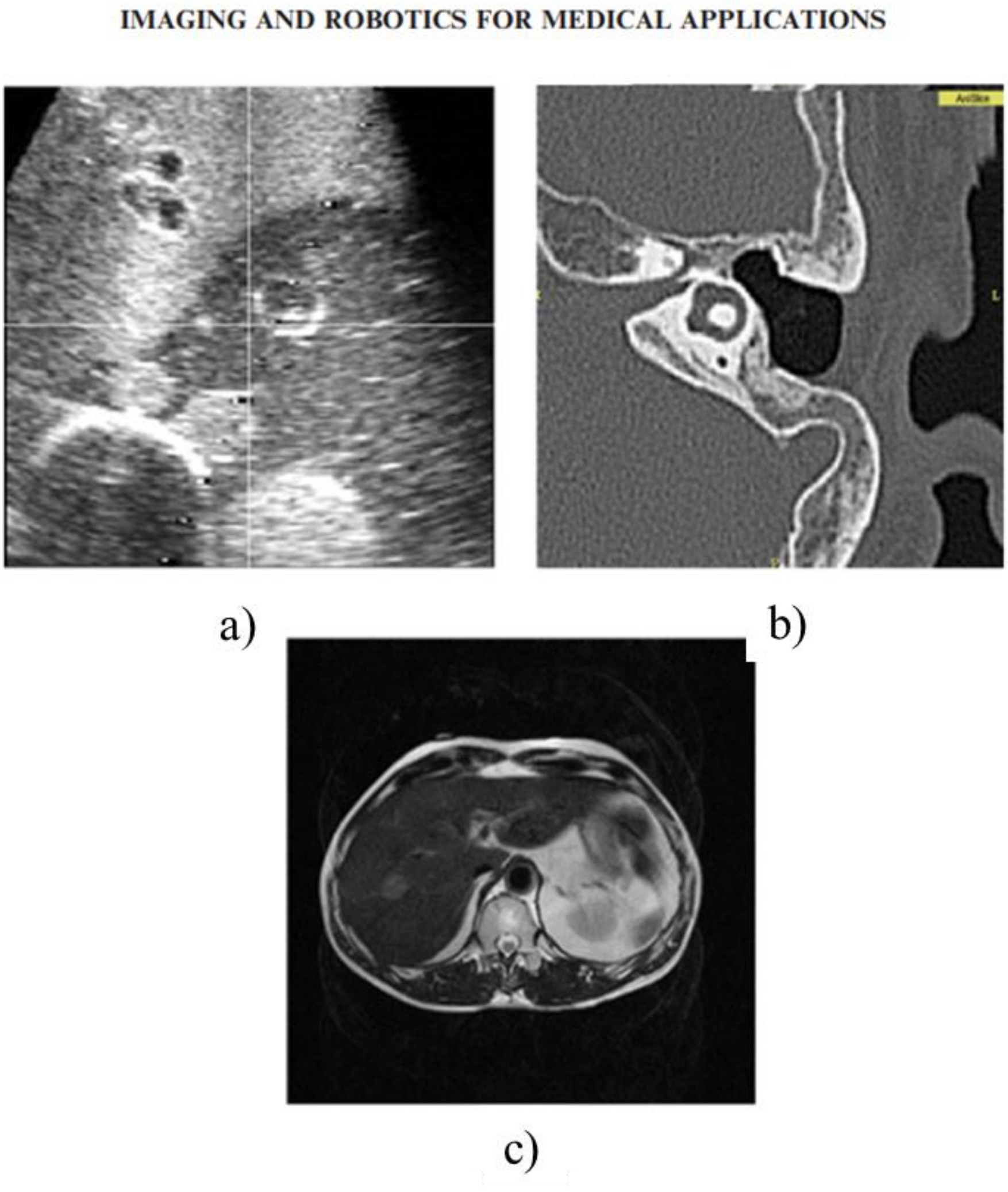

Figure 2. Comparison between US (a), CT (b), and MRI (c) imaging. Reproduced with permission.[65] Copyright 2011, Taylor & Francis.

Medical US, which is based on the amplitude and elapsed time of reflected sound waves, is normally implemented to reconstruct 2D images of the human organs and then provide valuable information to assist the surgical planning and execution during MIS procedures [66].

One of the main advantages of this technology is its safety, with low cost and portability. This technology has been widely installed in many hospitals and clinics worldwide as it offers the ability to provide real-time imaging beneath the surface of skin that is challenging to achieve from CT or MRI method [67]. To increase the view angle with minimization of hand tremor, a combination of endoscopy and ultrasonography is usually preferred [64-65]. Medical imaging with US is also used to assist with needle insertion therapies for the treatment of cancerous tumours with heat or ethanol [68]. MR and CT images might be also combined to precisely detect tumors and guide needle insertion [69]. For example, the Flex Focus 800 US system is an advanced imaging system, which offers precise real-time imaging on a high resolution 19-inch monitor display for intraoperative laparoscopic robotic surgery. It enables the surgeon to recognize and decide on the tumor targets that are unclear or have indistinct features [70]. The ACUSON X600 Ultrasound System which is manufactured by Siemens can deliver high imaging quality and reliability in a wide range of sophisticated applications. It leverages features of automatic tissue grayscale optimization and wireless data transfer to streamline operation across daily and shared-service imaging [71].

CT and MRI, on the other hand, are popular in diagnostic cross-sectional imaging because they provide better image resolution compared to US imaging [72]. Although CT images hard tissue like bone well, MRI displays images of soft tissue such as muscle and cartilage better [73]. However, radiation exposure is a drawback of CT and therefore this type of imaging is not ideal to perform repeatedly on the patient during surgery. In contrast, intraoperative MRI system has been proven to be a safe method for surgery. To create a 3D dynamic map of the surgical scenario, a combination of preoperative dynamic MRI/CT data and intraoperative ultrasound images has been carried out successfully [74]. The Aquilion Precision computed tomography system is the world's first ultra-high resolution CT. It can detect anatomy with sizes around 150 microns with technology rebuilding the CT images with improved high contrast spatial resolution. This system utilizes a deep learning algorithm to differentiate signal from noise and enhance the resolution of conventional CT two-fold [75]. The Philips Ingenia Elition solution offers cutting-edge MR imaging techniques with 3.0T imaging. It is able to reduce the MRI examination time by 50% and provides an immersive audio-visual experience to calm patients and to guide them through the MRI exams. This system can obtain a continuous and robust respiratory signal without any interaction and can automatically scan the target object via a pre-planning program [76].

Most robotic surgeries are reliant on a high-quality camera and video display to provide visual feedback to the surgeon during the operation [77]. Medical imaging technologies are also employed in endoscopy instruments[78]. A flexible endoscope is inserted into the human gastrointestinal (GI) tract with a high definition (HD) camera attached to its tip which transmits video to an external display screen located in the operating room. Near visual spectra is employed during the endoscopy in order to identify the GI abnormalities such as Barrett's esophagus [79]. Narrow-band imaging (NBI) can improve polyp assessment during colonoscopy procedures. Recent imaging technologies have been focused on new methods to enhance the angle view and to allow more precise tracking of the target. For example, the binocular eye-tracking system (Figure 3) is a retina-based eye tracker that can detect an eye gaze and measure depth perception and deformation of soft retina tissue [80-81]. Although endoscopy with advanced imaging systems have been used for many years, they are limited by the length of the scope which makes it difficult to assess the small bowel. Recent development in capsule endoscopy with a low image capture rate of around 2 Hz and wireless transmission of videos to external display device has enhanced the diagnosis of occult bleeding in the small bowel [82]. Better visual assistance and imaging guidance can significantly enhance the efficacy and safety of surgery, and decrease the operating times.

There are several medical imaging assisted surgical robots available in the literature. Readers can access to [83-84] for more details.

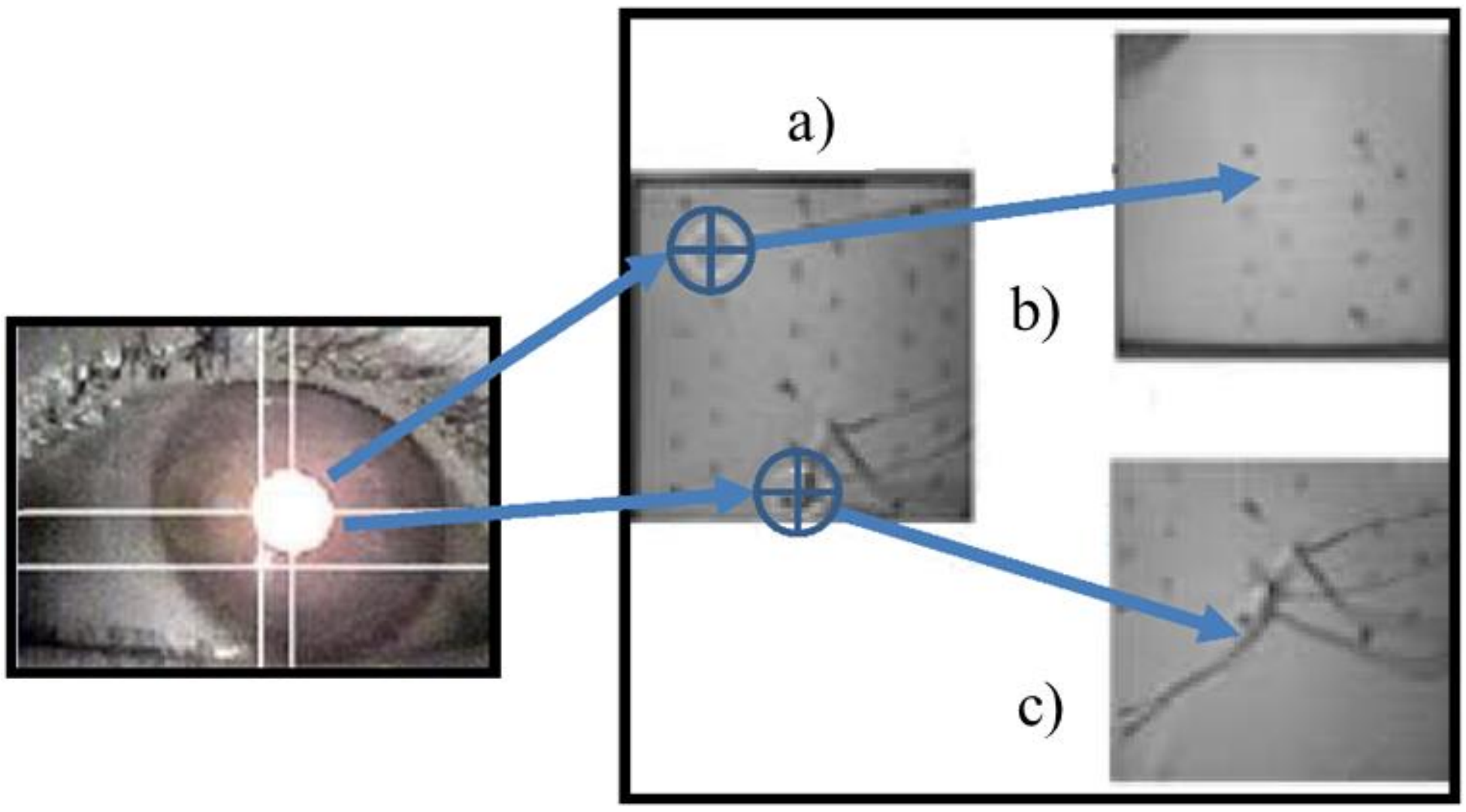

Figure 3. An eye gaze tracking integrated into surgical system. The eye gaze is focused at the top of the image (b). The gaze is focused to the bottom of the image; (c) The camera zooms to a defined area of the image. Reproduced with permission [81]

### 3.2. Force sensing for surgical robotics

Haptic feedback relates to the information and control associated with the human sense of touch [85-86]. The main disadvantage of present teleoperated robot-assisted MIS systems is the lack of haptic feedback to the surgeon about the interaction force between the tool tip and the target tissue. From the clinical perspective, the lack of feedback is a limiting factor for the surgeon to improve the accuracy and dexterity of surgery with the robotic system [87-88]. Without haptic feedback, the surgeon loses the ability to perform tissue palpation to assess tissue stiffness variations which can be performed during open surgery. Haptic feedback transmits the force/torque reading from the slave side to the master side via a haptic device that can reproduce the force interaction between the surgical tool and human tissue [89]. Surgical systems without the force feedback are not able to assist the surgeon to manage completely the tool-tissue interaction forces during an operation, especially in complex surgical tasks such as tissue suturing or knot-tying that requires more precision [90]. To overcome the problems associated with this lack of feedback, various sensing techniques have been developed to detect tissue interaction forces and to transfer the force sensing information to the surgeon [91]. Ideally, these sterilizable and biocompatible sensors are integrated at or near the tip of the surgical tool.

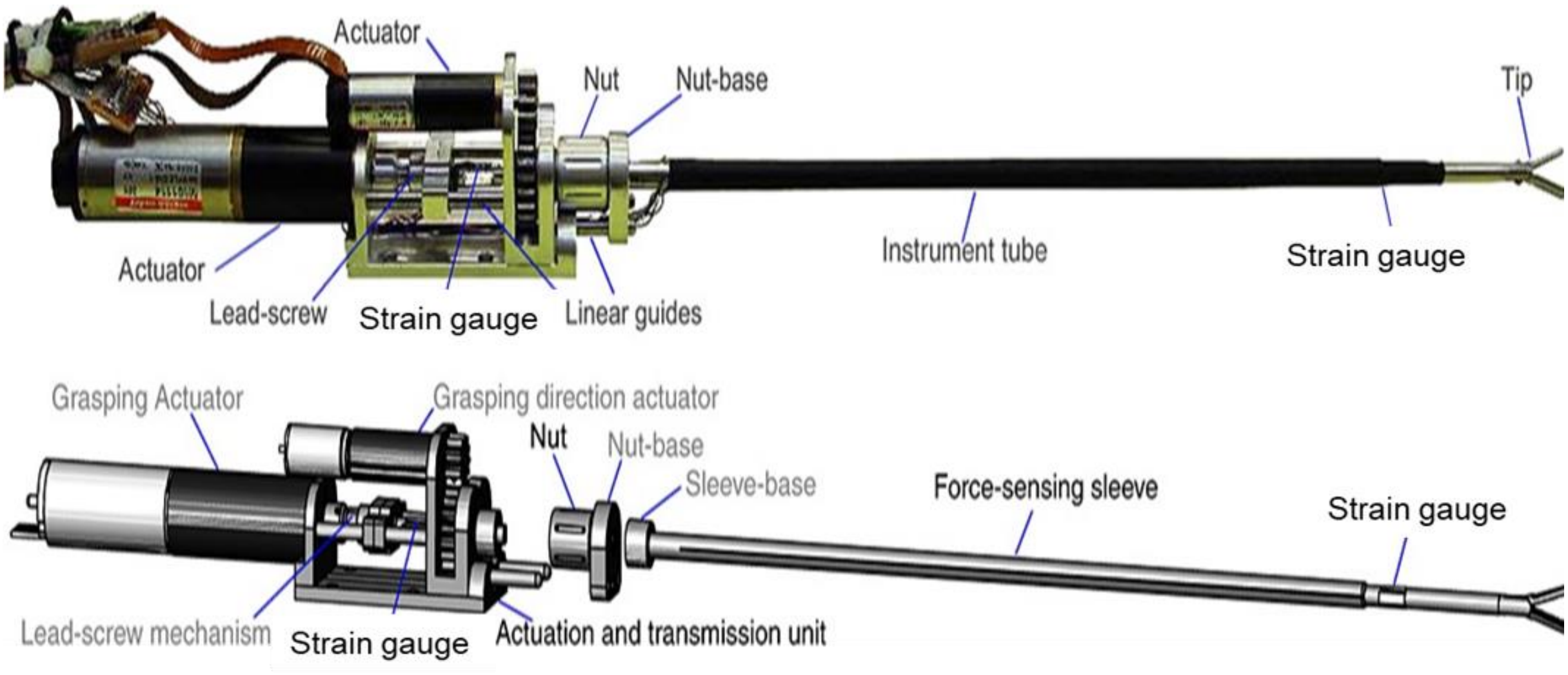

Figure 4. Prototype (top) and exploded 3D model (bottom) of the proposed force feedback enabled minimally invasive surgery instrument. Reproduced with permission.[105] Copyright 2013, John Wiley and Sons.

Strain gauges, capacitive sensors, piezoelectric sensors, and optical sensors are the most common force-sensing instruments in MIS [92]. Their materials typically consist of conductive inks, soft silicone elastomers, and dielectric structures [93-96]. Strain gauges which comprise thin metal foils are able to measure the interaction force for surgical tools based on a deformation of the thin film transduced to a change of resistance. Although a basic strain gauge is only able to sense one direction force, special arrangements of sensor cells at different locations and orientations or using novel compliant structures can provide multi-axis sensing measurements [97-98]. The strain gauge can be designed on a small scale to integrate into miniature surgical tools and devices with waterproof capability. However, this type of sensor can be affected by electromagnetic noise and temperature changes that may lead to inaccurate results [99-104]. Figure 4 illutrates a surgical instruments with integrated force feedback for MIS [105]

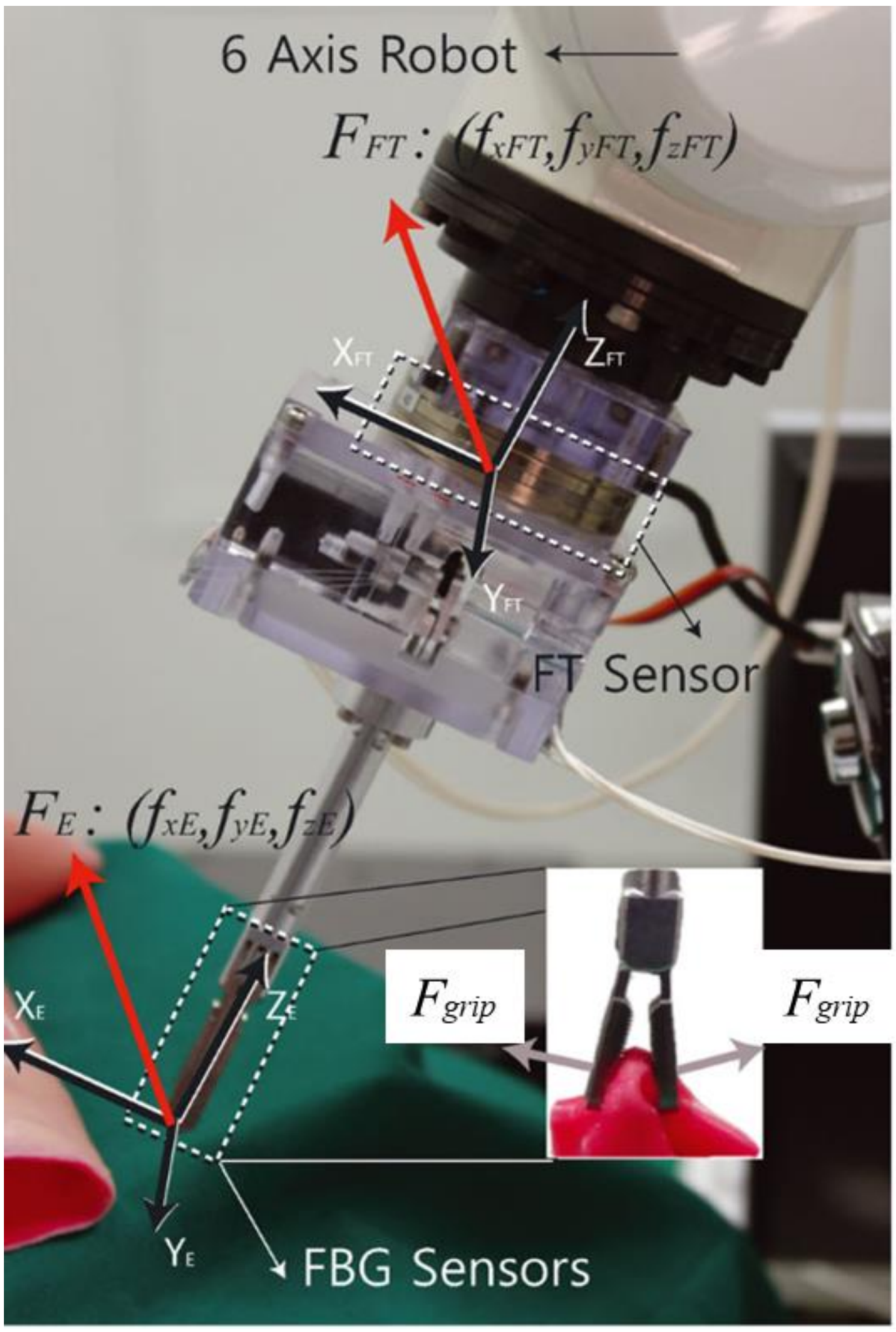

Figure 5. A 6-DOF serial robot, a force-torque (FT) sensor, a grasper, and optical fiber Bragg grating (FBG) sensor. Reproduced with permission.[120] Copyright 2014, John Wiley and Sons.

In contrast with strain gauges, capacitive sensors are able to provide force information based on the change of capacitance [106-109]. The sensitivity of this type of sensor highly depends on the Young's modulus of the dielectric layer which is sandwiched in between two electrodes. Once this layer is deformed under the applied pressure, the distance between the two electrodes or the thickness of the dielectric layer is varied. A signal processing circuit will convert the capacitance change to the applied normal force or shear force via mathematical models [110]. Such type of sensor is simple and easily reproducible, which is suitable for disposable use [111]. In addition, it also offers better stability and sensitivity in warm and wet environments compared to strain gauges. However, capacitance sensor requires more intricate signal processing and special sealing method that are normally associated with a complex process. Piezo materials are used for both sensing and actuation purposes based on voltage changes in piezo plates that are related to changes in mechanical stress and strain [112-114]. Their additional advantages include high bandwidth, compact size, and high-power density. Despite advances, they are also limited in dynamic loads and are affected by temperature changes from surrounding environments [115-117].

Optical sensing, on the other hand, provides force measurement of up to six DOFs based on the changes of intensity or phase of light passing through a flexible tube to a compliant structure [118]. This type of sensor can work under a magnetic environment with less hysteresis and reproducibility. However, its sensitivity depends highly on the flexible tube materials and the alignment of the compliant structure. In addition, it can only detect large bending radii that may prevent its use in miniature devices [119]. Figure 5 demonstrates a 6-DOF surgical tool with the integration of a force/torque (FT) sensor, a grasper, and optical fiber Bragg grating (FBG) sensors [118].

Other force measuring technologies for surgical robotic systems are available [91, 103, 107, 121-125]. Hybrid configurations between two or more force sensing technologies are available. For example, biofeedback sensors can be integrated into surgical tools to monitor the oxygen saturation levels of the tissue and alert a surgeon if the oxygen level is below an acceptable level while strain gauges can be used to measure the applied force of surgical tools [126]. US transducers can be used to provide the interaction force information based on the deformation of the tissue with respect to the US pulse traveling time [68, 127-129].

### 3.3. Position sensing for surgical robotics

Most surgical robotic devices function by commands from the surgeon's hands and eyes. One major challenge for surgical devices is the lack of real-time position feedback when the surgical tool is manueuvred inside the human body. Position feedback including the relative position of the robotic joints and the curvature of flexible parts play a vital role in enhancing accurate system performance [130]. Recently, approaches to real-time position sensing have been studied using imaging, light, electromagnetic tracking and onboard sensors such as stretchable piezoresistive/capacitive sensors [131-132].

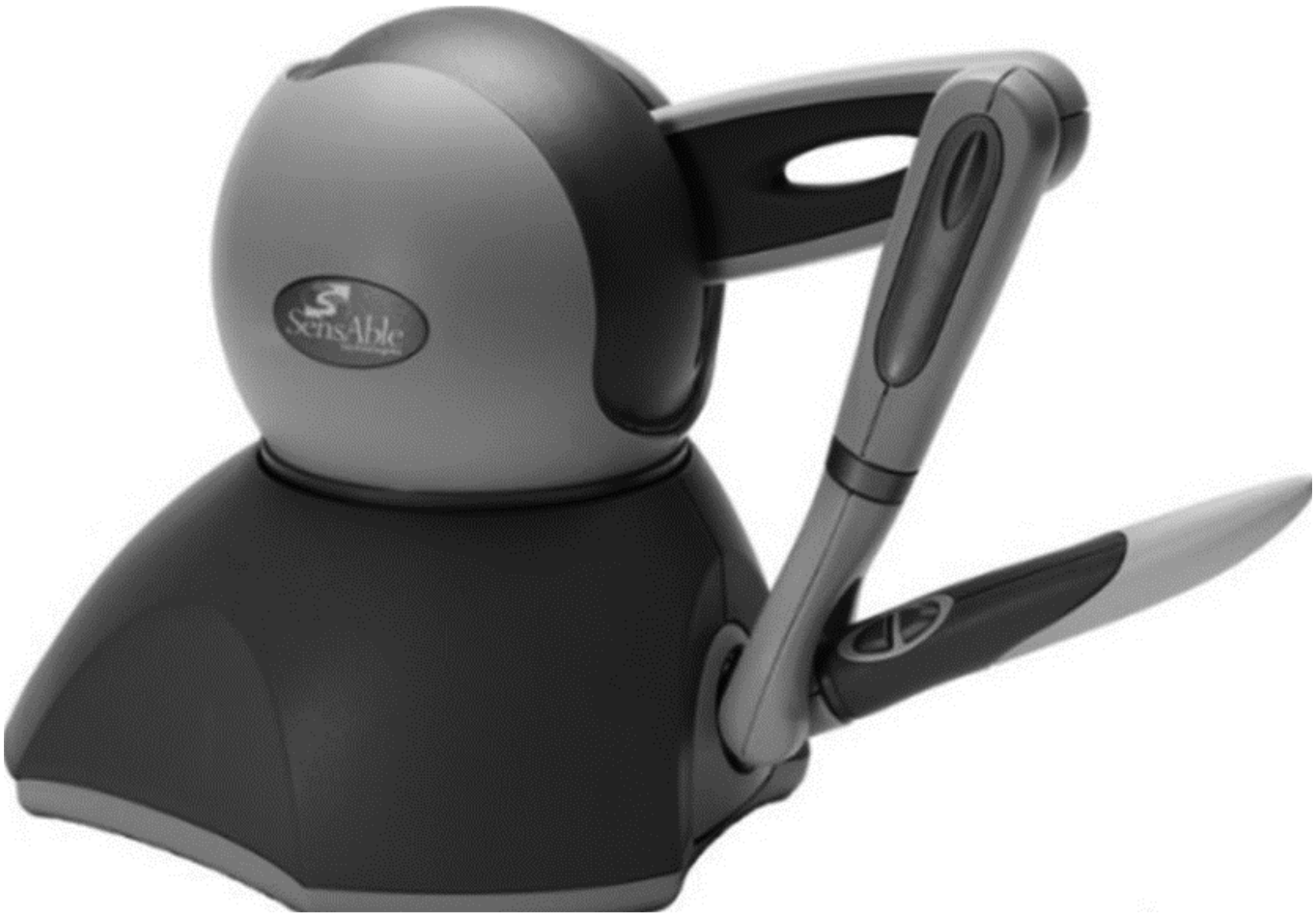

Figure 6. Phantom Omni. Reproduced with permission.[136] Copyright 2015, John Wiley and Sons.

In existing surgical systems, manipulators are usually driven by DC motors equipped with encoders to provide position feedback. The DC motor can transmit rotations or translation to the surgical tools via flexible cables or mechanical linkages [19]. Most surgical system operate in open-loop control where the real-time position of the robotic joints are absent, only position information from external encoders are available [133]. The main controller processes the absolute encoder data to estimate the position of robotic joints inside human's body and then use it as a feedback information to detect over-speed or to warn any undesirable collisions between the human's organs and surgical tools [134]. This estimated feedback normally ignores nonlinearities such as backlash or hysteresis of the transmission system where precise positions are not strictly required such as applications used with the SensAble™ PHANTOM Omni (SensAble Technologies, Inc., Woburn, MA) (Figure 6) [135-136].

Recently, many surgical systems have used advanced position sensing technologies to enhance their performances during operations. Celera Motion has developed a series of custom-sized PCB-based encoder modules with a standard 40-micron pitch incremental counting track, Fine Ball Grid Array (FPGA) control, and Bidirectional interface for Serial/Synchronous (BiSS) communication interfaces. This system has been successfully implemented into radiotherapy cancer treatments with high speed and high-resolution of beam collimation [137]. Tekscan has developed a thin potentiometer for position feedback in surgical tools, namely the FlexiPot™ Strip and Ring tactile potentiometer [138]. These sensors can be used to identify and adjust the location or position of the applied contact force to the surgical tools. To sense the surrounding objects, a whisker sensor was developed to estimate the relative position of robotic arms with respect to nearby organs [130]. The first version of this sensor used two flexible cantilever beams to measure linear motion along the tip while other two-dimensional lateral motions are determined by strain gauge sensors. The latest version of this sensor was equipped with a cross-shaped flexible structure that is located at the back of the linear sensor, enabling measurement capability. The main advantages of this cross structure include higher stiffness and smaller dimensions with no friction and backlash. However, limitation on the deflection is a drawback of this design [130]. Recent developments in soft robotics enable advanced technologies for position sensing in surgical robotic system [132, 139-143].

Optical tracking systems and electromagnetic tracking systems are other non-touch sensing technologies that are employed in surgical robotic systems [144-145]. To provide the position feedback, pointed tips or markers are usually attached to the surgical instruments such as electrocautery devices, surgical drills, and endoscopes. In optical tracking systems, cameras are normally used to provide position feedback using image processing to detect the position of visual markers or the reflected signals from infrared light or the visible spectrum. Although imaging tracking methods have shown to have high accuracy, the obtained signals are highly dependant on the visualization of the detected objects or markers that sometimes are obstructed by internal organs or surgical tools [146]. Therefore, optimal positions of the camera and markers are strictly required [147]. A recent study showed that a hybrid marker design from circular dots and chessboard vertices (Figure 7) can significantly enhance the tracking result [148]. In contrast to the above approaches, Fiber Bragg Gratings (FBGs) have received attention from research communities due to their small size, biocompatibility, and high sensitivity [122]. This technology is the best suited to detect the 3D bending motions of flexible robotic arms within complex channels such as blood vessels or the human GI tract [149]. For electromagnetic tracking systems, markers are helical coils that are integrated into the surgical tool tip to provide the position and orientation with respect to a global coordinate. The position and orientation of the tool is determined via the change of an electric current sequentially passed through the coils under the external magnetic field [150]. This method provides good accuracy even in cramped spaces with no light. However, the magnetic fields can be affected by external environments such as surrounding ferrous materials. To enhance the sensor accuracy, a combination of preoperative models and intraoperative tracking data is recommended [151-154]. Other tracking methods proposed in the literature include the estimation of articulated tools in 3D using CAD models of the tools [155].

### 3.4. Advanced actuation for surgical robotics

To drive the surgical tools, actuators are remotely installed away from the end effector and surgical sites, offering a simple and safe solution for a light weight instrument to operate inside the human body [19]. From the clinical needs of smaller size and higher flexibility for surgical tools, advanced actuations including cable-driven mechanisms, flexible fluidic actuators, smart material actuators, and magnetic actuators have been developed to transmit the force and motion from the actuator site to the end effectors [156]. Although surgical robotics have been used over the past few decades, there are many intrinsic difficulties that constrain their use in some surgical procedures where miniature size and high applied force of surgical tools are highly desired. This section will overview advanced actuation methods that are mainly used in existing surgical robotic systems.

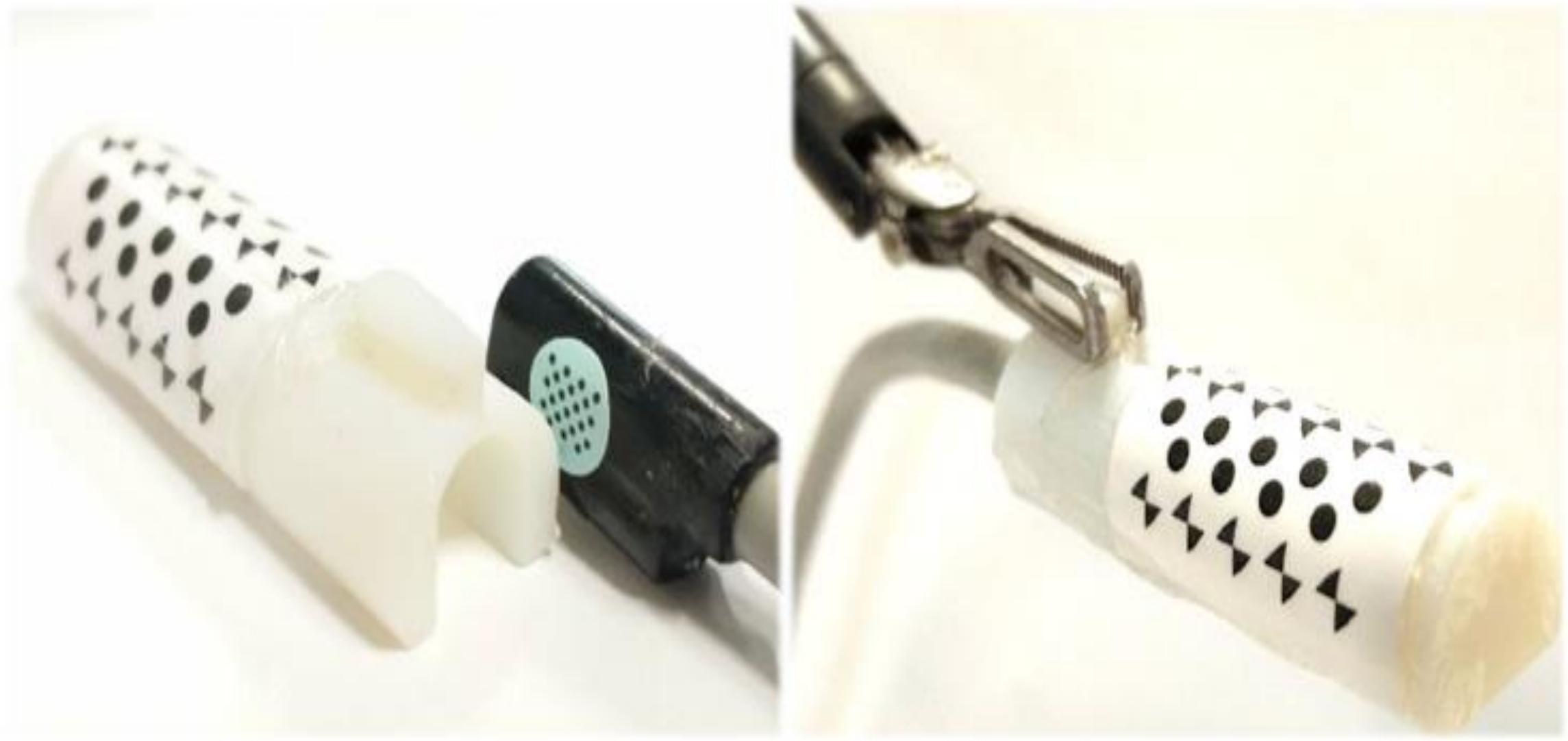

Figure 7. A housing adaptor that can hold a planar ultrasound probe. Reproduced with permission.[148] Copyright 2017, Springer Nature.

Cable-driven actuation has been proven to be an important transmission mechanism in many surgical systems including the commercial Da Vinci surgical system and Medrobotics Flex system [157]. This type of transmission system overcomes major drawbacks in conventional onboard DC motors that are normally associated with bulky size and heavy weight [158]. In cable-driven mechanisms, desired position or force are transferred from external actuators to distant joints via fixed points (pulleys) or routing inside flexible tubes (sheaths). Figure 8 shows the difference between the cable-pulley and the tendon–sheath actuation. Many studies have shown that the tendon–sheath mechanisms possess better capability to operate in unpredictable channels and confined space. The tendon-sheath configuration has been widely employed in many surgical robotic applications and flexible surgical devices, especially in robotic catheter and flexible endoscopic systems [159]. Despite advances, high force loss due to nonlinear friction and backlash hysteresis between the cable and the outer sheath degrades the system performances. Figure 9 shows an early version of the MASTER surgical system with two robotic arms ( nine DOFs for each) driven by tendon–sheath mechanisms. In contrast, the cable with pulley configuration offers higher force transmission due to minimized friction effects and less nonlinear hysteresis [160-161]. Most laparoscopic surgical systems use the cable-pulley system as the main mode of transmission. For example, in the Da Vinci Surgical System, most disks or vertebrae are stacked together to build the wrist structure. To control the tool motion, cables are connected to the distal joints via attached pulleys and then transmit the motion from the proximal vertebra to the intermediate or to the distal vertebra. To mitigate the nonlinear effects such as backlash and hysteresis, initial pretension is applied to the cables, preventing them from being slack [52]. In endoscopic systems or robotic catheters, a hybrid combination between flexible cables and sheaths is preferred to control the bending motion of the flexible part via the incompressible and ring-shaped elements that are connected by hinges, in a manner similar to the human spine (Figure 10). With this configuration, the cables slide over tiny sleeves to control the bending motion of the tip in two directions (except the axial direction (Figure 11)) [162]. In some cases, spine mechanisms are simplified to a flexible tube with notches instead of hinges, making the flexible part stiffer. Cables were also combined with cylindrical concentric tubes to provide a highly articulated robotic probe (HARP), that can offer three-dimensional arc commands [163]. The Medrobotics Flex System adopts this design.

Flexible fluidic actuation is also used for medical applications [164-165]. This type of actuation converts the pressure from an external fluid source (hydraulic or pneumatic source) into elongation or bending motion based on deformation of elastic materials, which subsequently actuates the robotic joint or flexible bending part [166]. Fluidic actuation can bend, stretch or rotate surgical tools with several DOFs [167]. Most of the bending motions induced by fluidic actuators are a type of anisotropic rigidity where the internal chamber is sandwiched in between two soft different stiffness layers. Under applied pressure, the length or surface area of stiff layer expands less than the softer layer, inducing a bending motion towards the stiffer part. This phenomenon can be seen in Figure 12 while Figure 13 shows soft balloons-actuated micro-fingers with two silicone parts made from a parylene balloon. An elongation actuator by a fluidic source typically consists of an elongated silicone part with an internal hollow chamber and a constraint layer wrapped around its circumference made from inextensible fiber. Under applied pressure, the silicone part will be axially elongated without any radial expansion due to the constraint from the inextensible layer [168]. Detailed discussion for this type of actuations can be found in a recent study by Minh et al. [19].

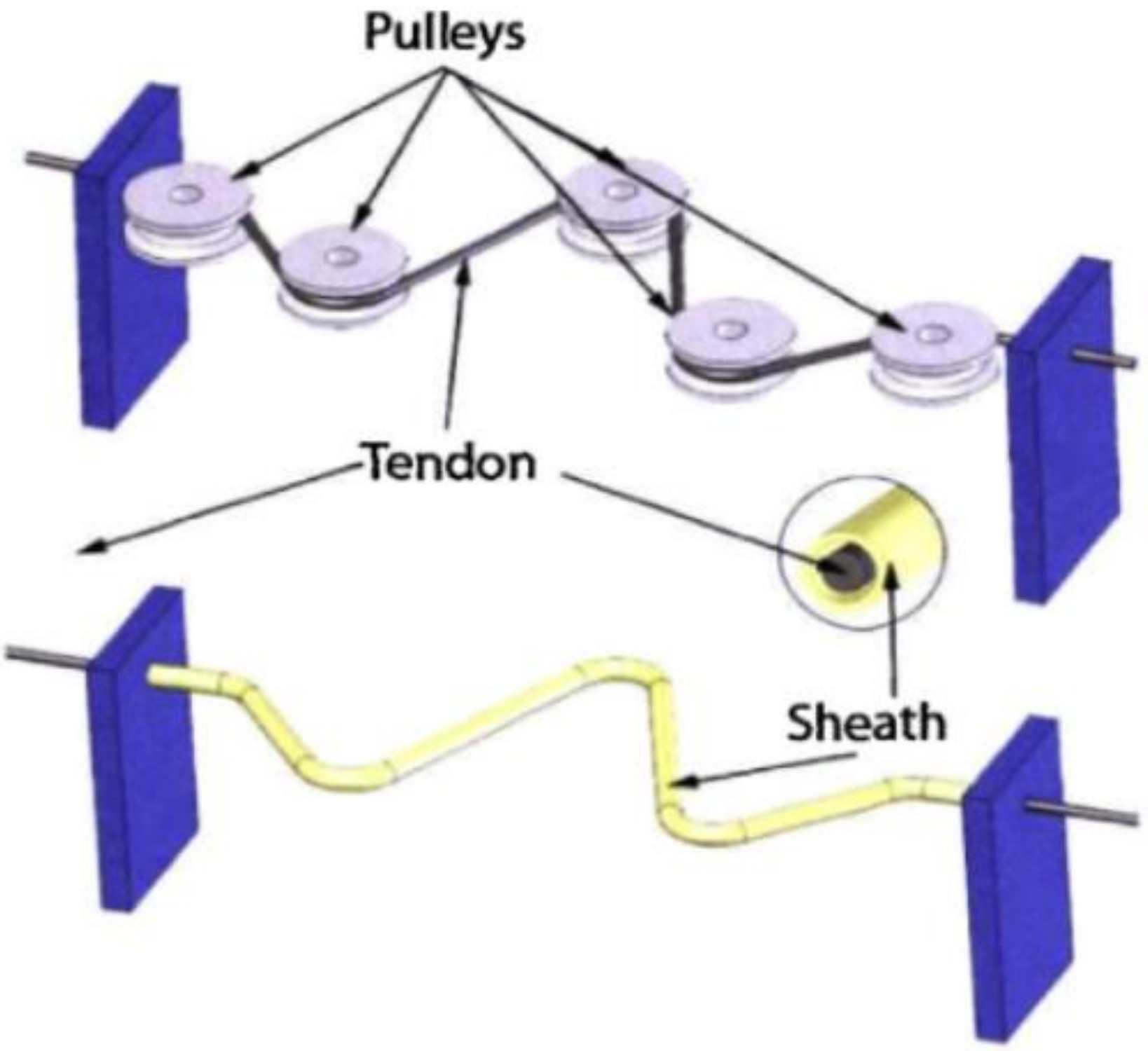

Figure 8. Difference between the cable-pulley and the tendon–sheath actuation. Reproduced with permission.[156] Copyright 2016, Elsevier.

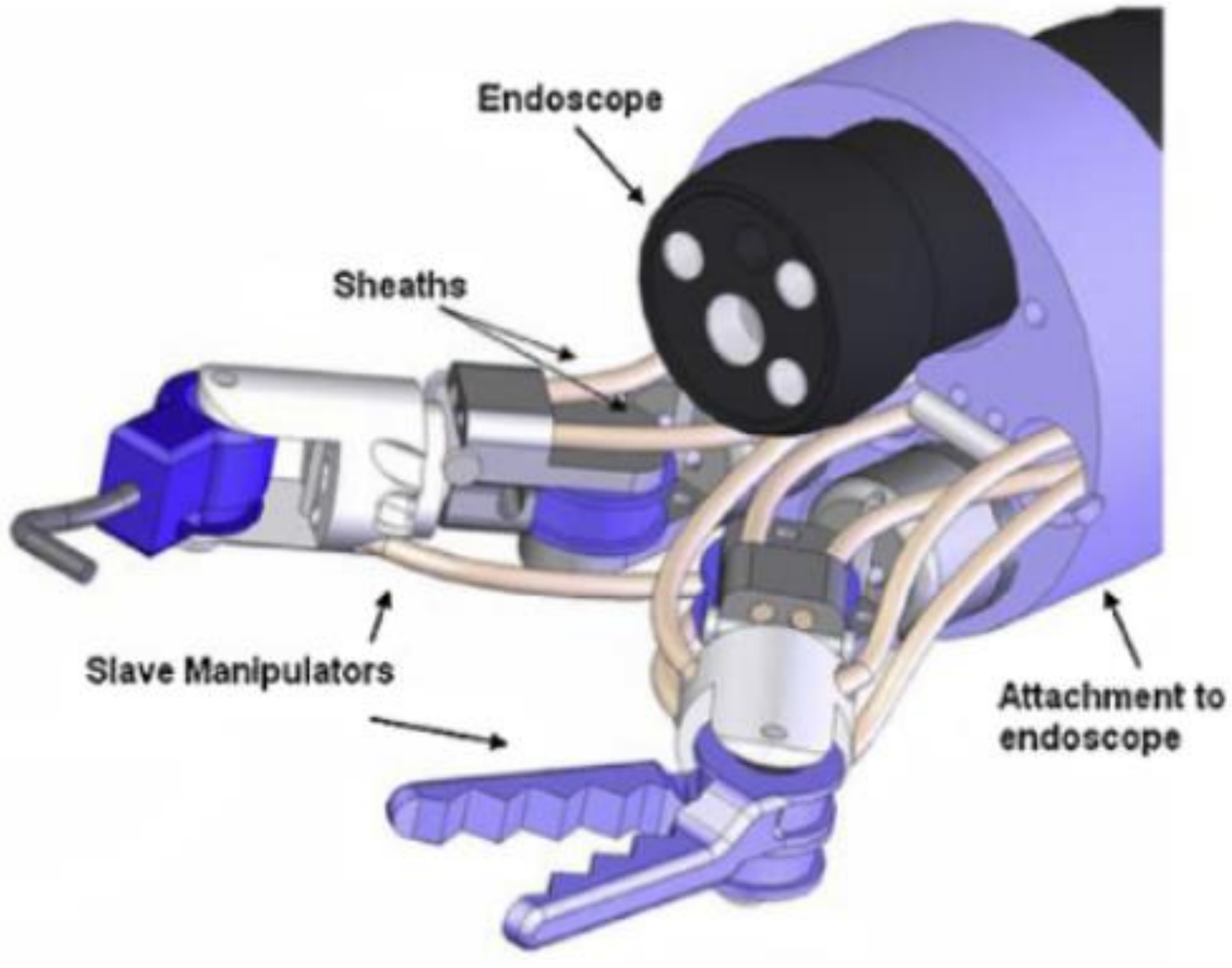

Figure 9. The slave manipulators with nine degrees of freedom. Reproduced with permission.[156] Copyright 2016, Elsevier.

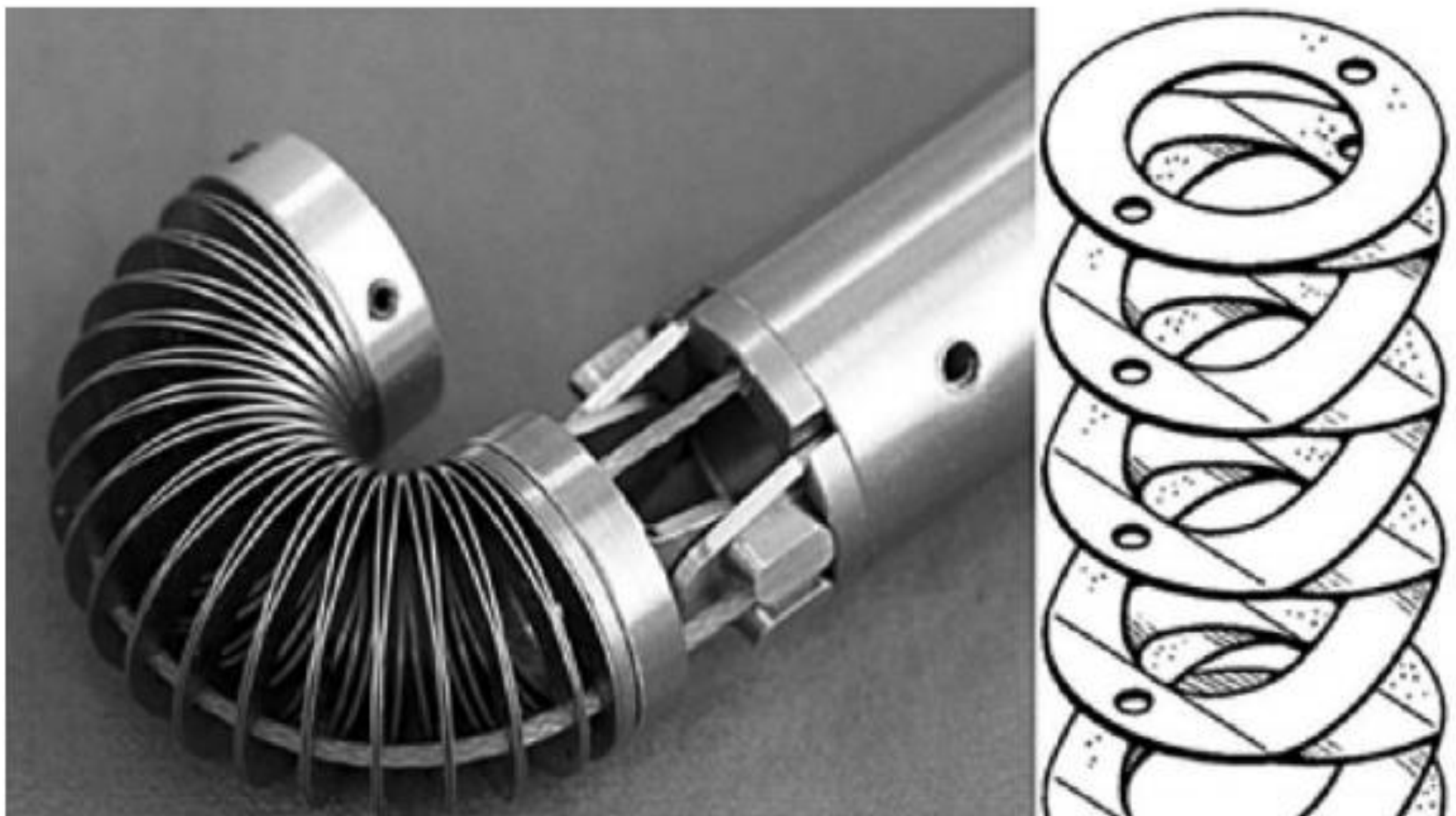

Figure 10. Steerable tip and ring-spring version I with diameter of 15 mm. Reproduced with permission.[156] Copyright 2016, Elsevier.

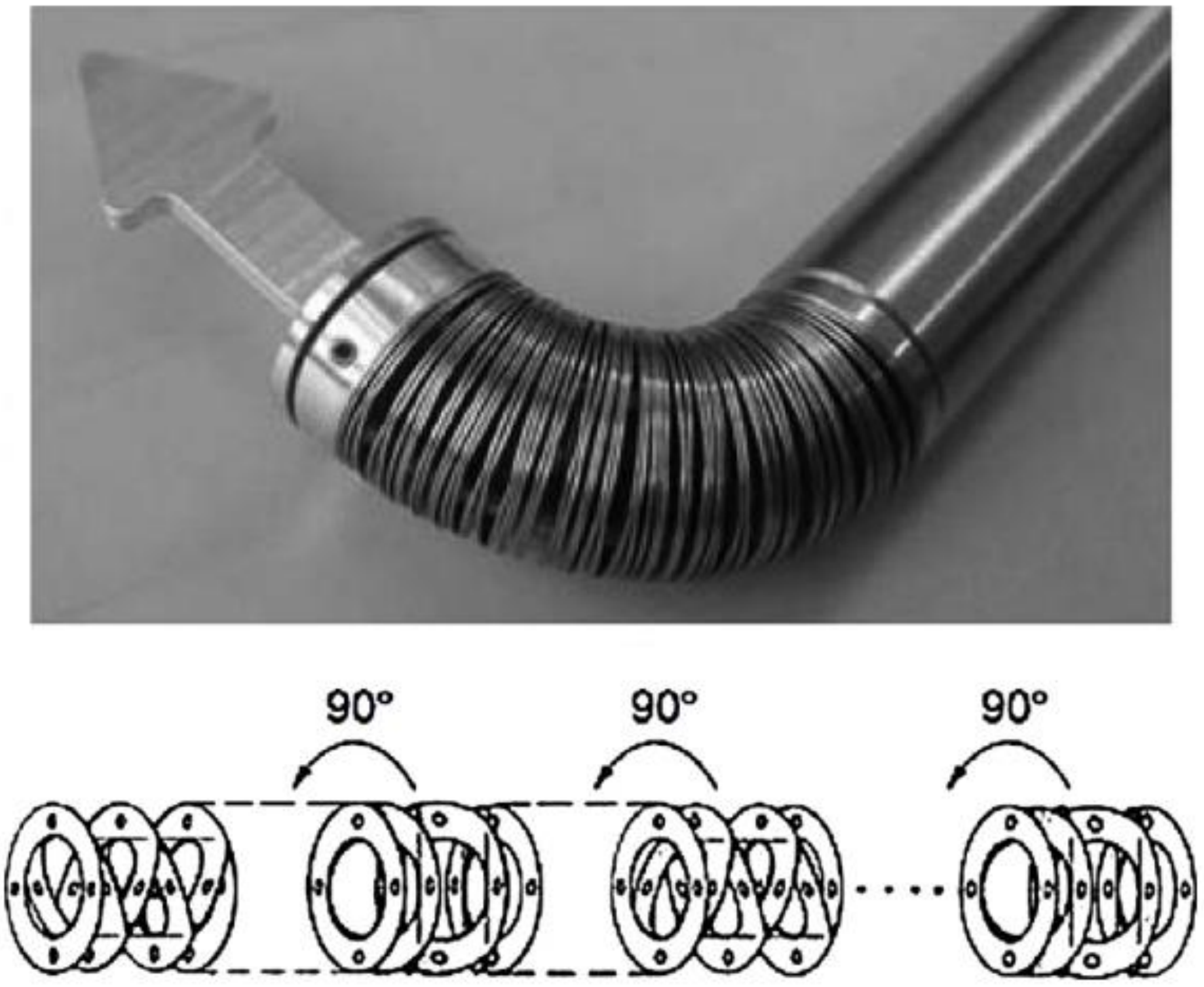

Figure 11. Steerable tip and ring-spring version II with a diameter of 12 mm. Reproduced with permission.[156] Copyright 2016, Elsevier.

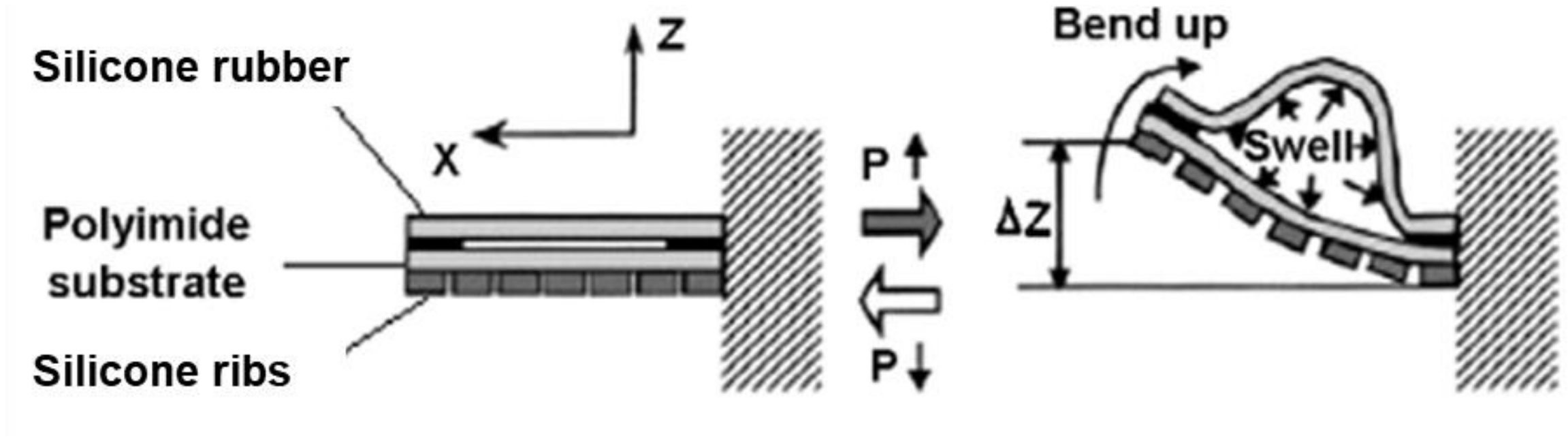

Figure 12. Flexible fluidic actuators. Reproduced with permission.[164] Copyright 2009, Elsevier.

Shape memory alloys (SMAs) and piezoelectric actuators are other types of actuation with smart materials, which have been recently implemented in surgical robotic systems [169]. SMAs are able to memorize and recover to their previous shapes based on the change of temperature or magnetic stimulus [170]. Due to the high work density and biocompatibility, this type of actuation has been used in many surgical robotic systems such as robotic catheters, endoscopes, and surgical graspers [169, 171]. Figure 14 introduces an SMA spring coil covered by a soft silicone tube that can change its longitudinal shape under applied heat. Ultrasonic actuators that are constructed from piezoelectric materials can generate higher torque, low electromagnetic radiation, and precise motions using the voltage as the control input. The ultrasonic motor has the capability of generating multi-DOF rotation as a spherical rotor (Figure 15), which is comparable to the human wrist motion. This type of actuator has been used to control surgical forceps with multi-DOFs [172]. For linear actuation, the piezoelectric actuator has been incorporated into biopsy probes such as the ROBOCAST system, surgical tools, and endoscopes [173]. This type of actuator is also compatible with MRI and therefore can be used in surgical procedures where imaging feedback is available [174].

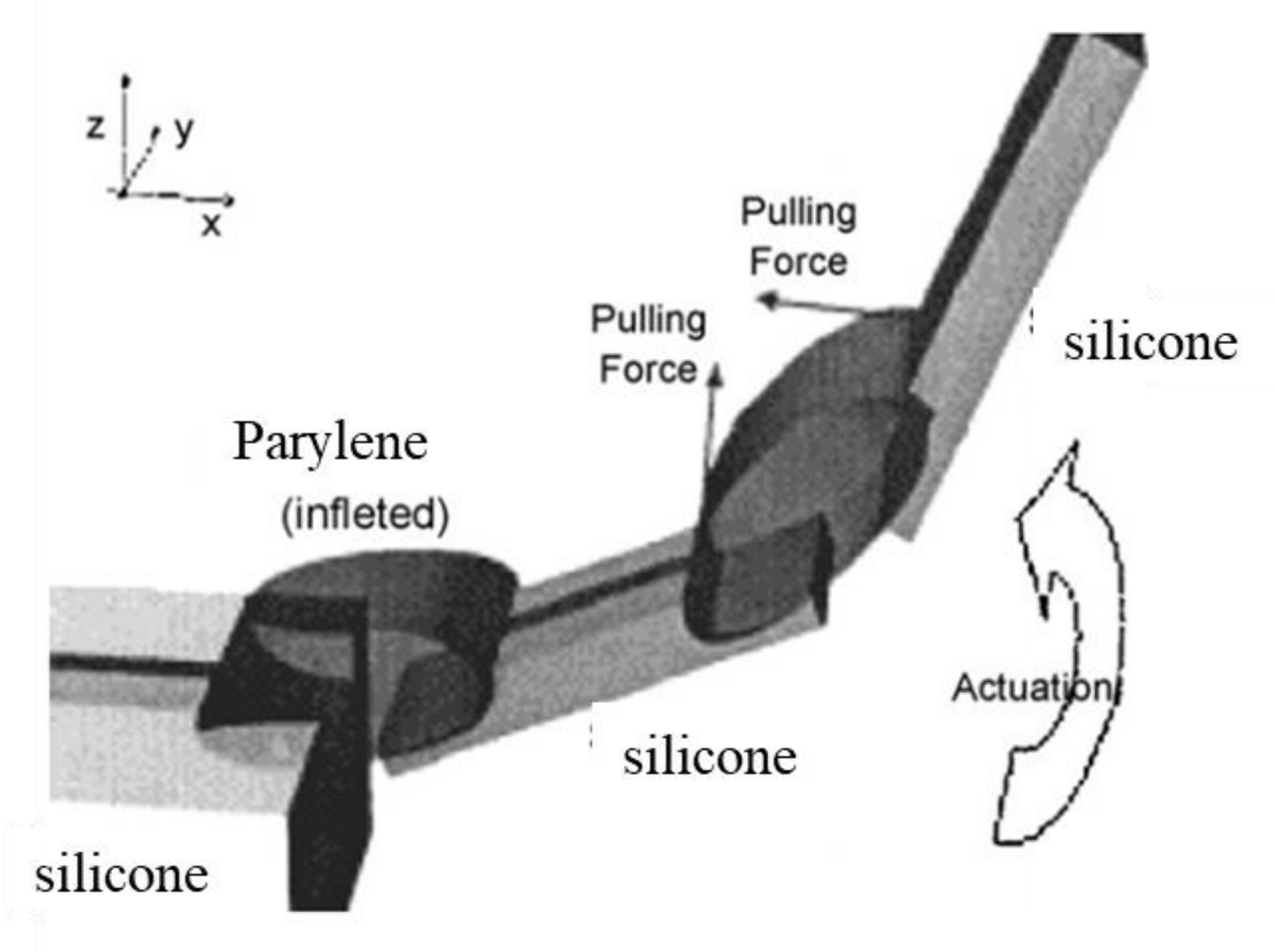

Figure 13. Soft balloons-actuated micro-finger. Reproduced with permission.[164] Copyright 2009, Elsevier.

Magnetic actuation, on the other hand, shows different characteristics compared to other types of actuation methods. In surgical applications, magnetic fields that can be generated by a permanent magnet or electromagnet are used to wirelessly control the surgical tools without requiring a physical transmission link between the power source and the tools [175-178]. To implement the magnetic actuation in surgical tools, magnetic generators are externally located outside the patient's abdominal wall where the actuation force/torque is magnetically transmitted from outside via the abdominal wall [179-181]. This means that on-board electromagnetic motors or cables are completely eliminated. The magnetic actuation has been widely applied in many surgical procedures. For example, the tissue retractor for liver resection consists of two pairs of magnets where one pair plays a role of anchoring while another pair connects to a retracting lever ( Figure 16) [182]. In another approach, a 4-DOF surgical manipulator for MIS was constructed to execute surgical tasks. This system combines local magnetic actuation and cable-driven actuation with an outer diameter of less than 15 mm to enable supportive forces against the tissue. Despite advances, this type of actuation has some disadvantages such as high nonlinear hysteresis. In addition, the interaction force is highly affected by external environments, especially when the surgery is operated near by ferromagnetic materials [183].

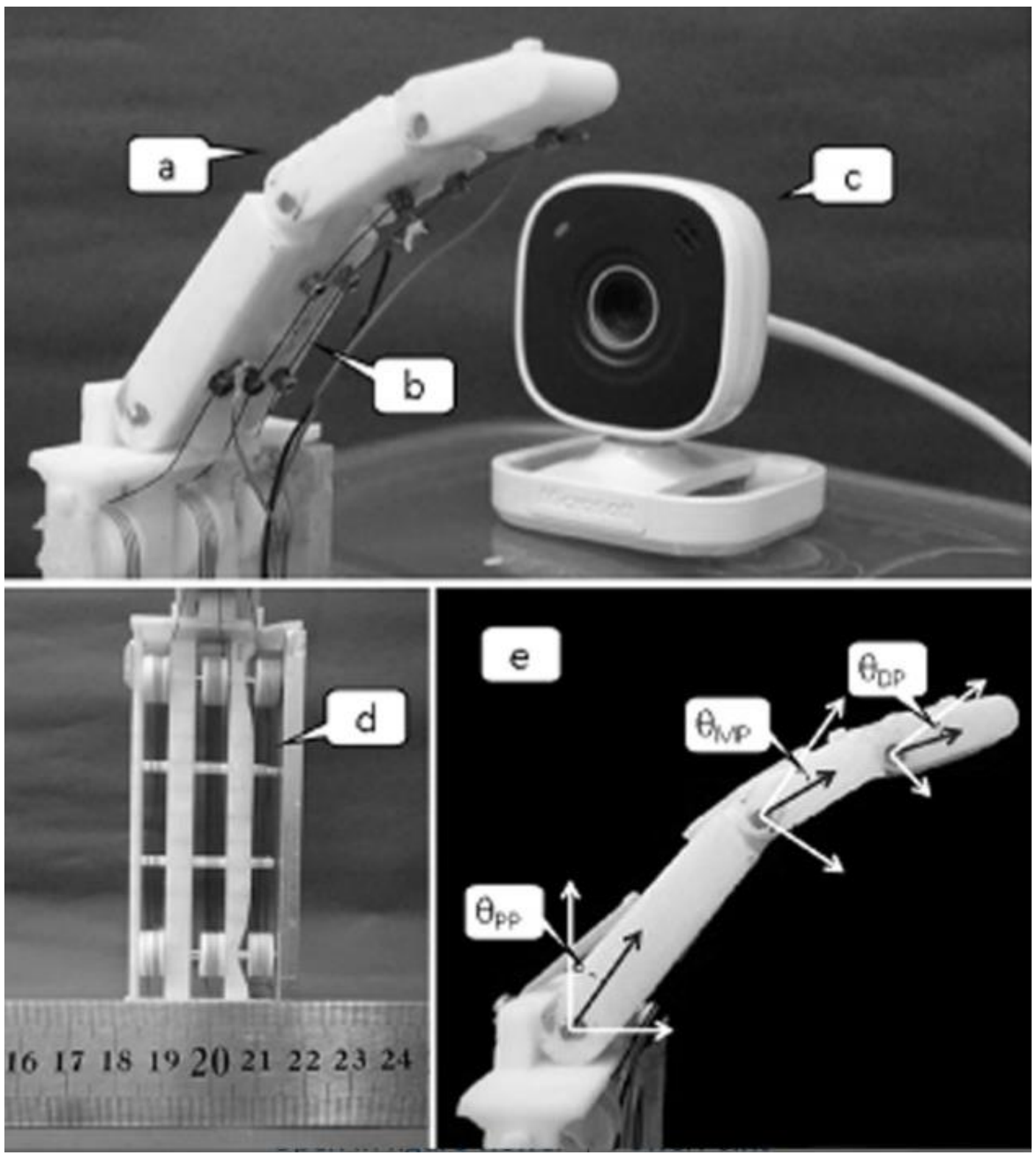

Figure 14. Photograph of the shape memory alloy (SMA) finger manufactured by rapid prototyping. Reproduced with permission.[328] Copyright 2013, John Wiley and Sons.

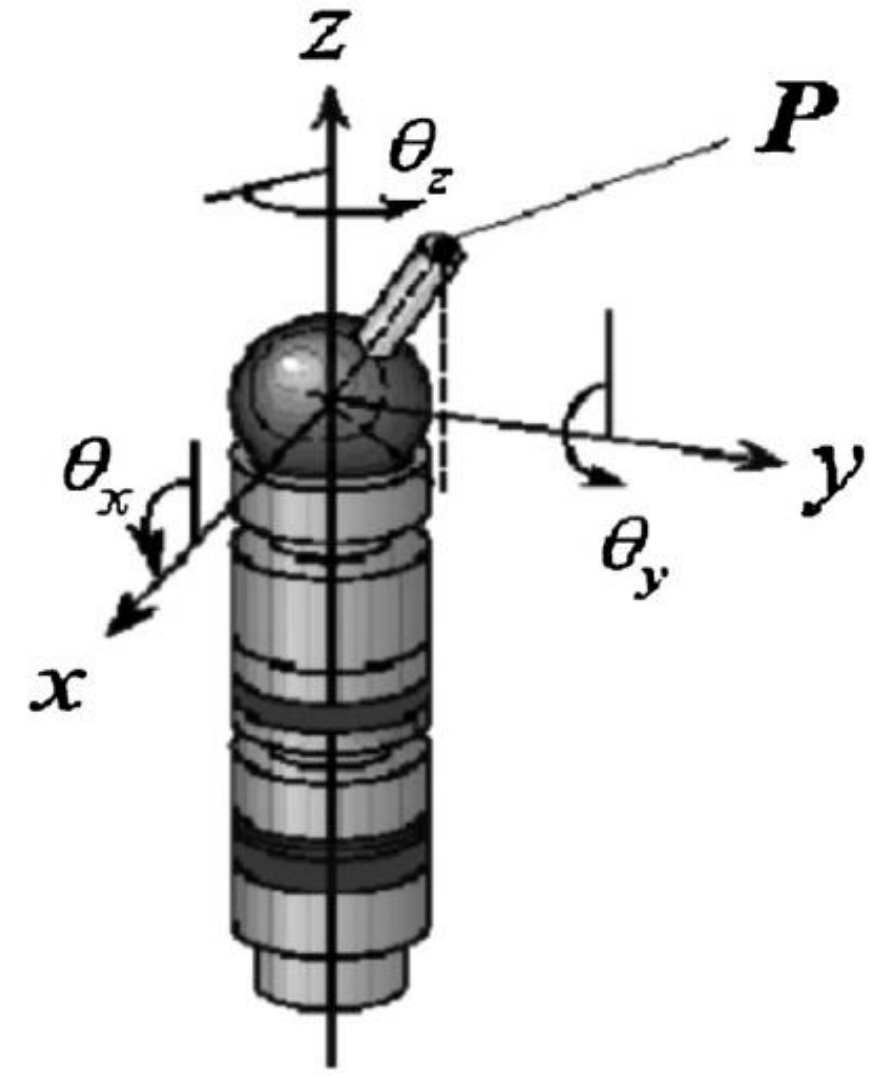

Figure 15. Multi-DOF ultrasonic motor. Reproduced with permission.[156] Copyright 2016, Elsevier.

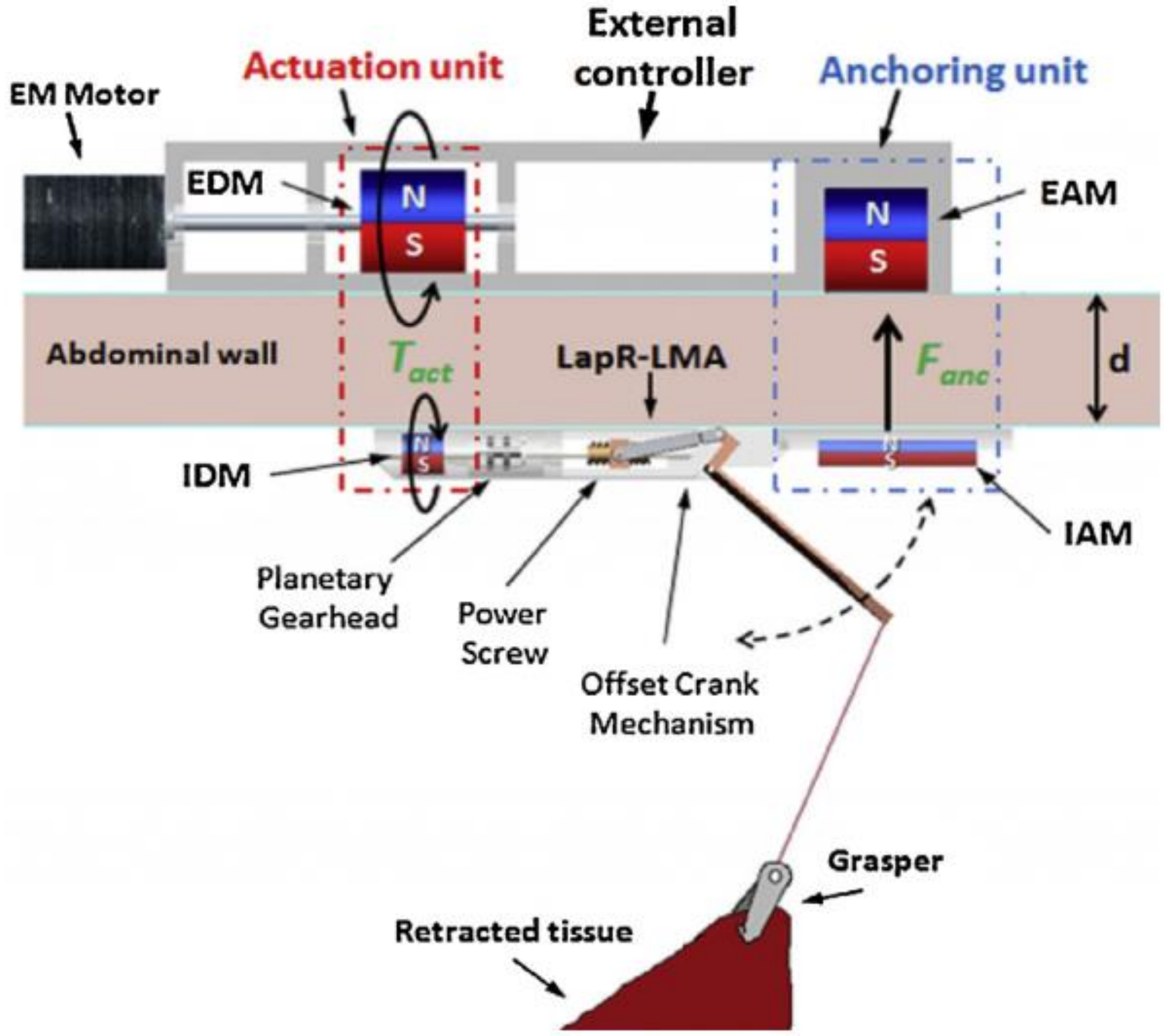

Figure 16. Schematic representation of the LapR-LMA using magnetic actuation. Reproduced with permission.[156] Copyright 2016, Elsevier.

### 3.5. Haptic display systems for surgical robotics

Haptic means pertaining to the sense of touch, which enables humans to grasp and manipulate objects, to feel and distinguish various surface properties of materials [184]. The sense of touch is distributed over the entire human body and is typically divided into two types: kinesthetic and tactile. Kinesthetic haptics include forces and torques that are sensed in the joints, muscles, and tendons while tactile haptics such as vibration, pressure, and shear force, are sensed by mechanoreceptors under the skin [59, 88, 185-187]. Most teleoperated surgical robots employ kinesthetic sensations as major haptic feedback because of its simplicity compared to tactile feedback. However, investigatin of tactile feedback, in the form of lateral skin deformation and vibration, has risen recently due to advantages relating to compactness, wearability, and cost-effectiveness [26, 188-189].

Haptic systems can be classified into three main categories [188]. First, graspable systems are normally kinesthetic devices with a handheld tool, allowing force interaction between devices and users. Second, wearable systems are typically tactile devices that can provide haptic information directly to the skin. Lastly, touchable systems are mixed-mode devices that offer users the ability to actively examine the properties of the entire surface. Haptic feedback remains one of the major challenges in the development of surgical robotic systems [61]. The lack of haptic information to the surgeon can lead to a lack of or excessive force application to the target during surgical procedures, resulting in slipping or damaging of tissue [190]. Haptic feedback can be used for surgical training and to improve the function of surgical instruments [191]. In this section, an overview of several devices and methods that integrate haptic feedback into medical applications are reviewed.

The Geomagic® haptic device (3D Systems Inc., Rock Hill, South California, USA) was developed for research, 3D modeling, and original equipment manufacturer (OEM) applications [192]. Model Touch™ and Touch X™ provide 3 DOFs force feedback and 6 DOFs navigation including 3D spatial positioning and 3D orientation (Figure 17) [193]. These devices offer a wide range of motion, analogous to the human hand pivoting at the wrist. Force feedback is generated by DC motors and is transferred to the users' hands via the handheld stylus. This stylus can be customized to imitate surgical instruments for use in medical simulation and training exercises.

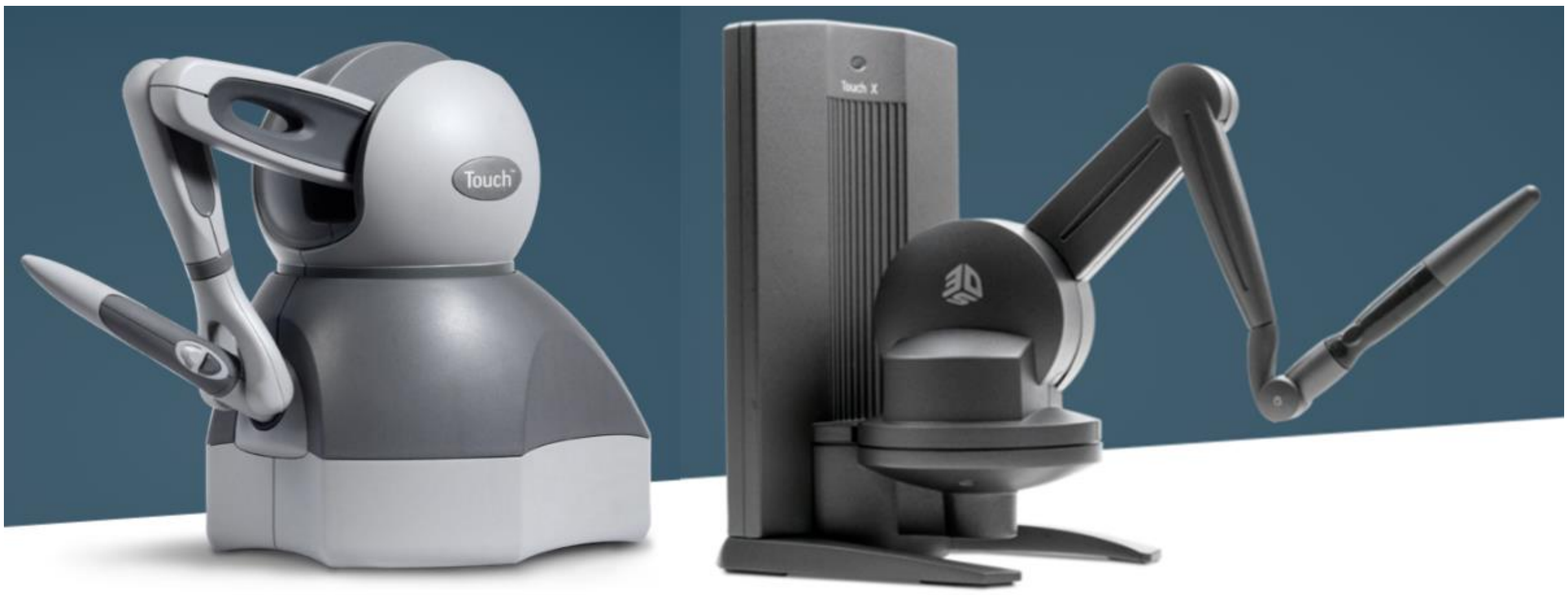

Figure 17. Touch™ (left) and Touch X™ (right) (© 3D Systems, Inc.)

There are a number of haptic devices that are mainly developed for surgical systems including the OMEGA Haptic device [194] (Figure 18). Pacchierotti et al. [58] introduced a haptic feedback system with a BioTac tactile sensor embedded into the distal end of a surgical tool and a custom cutaneous feedback device at the surgeon's fingertips. The BioTac sensor (Shadow Robot Company, UK) is responsible for sensing contact deformations and vibrations with the object and then transmitting these signals to the fingertip to induce the haptic feedback. Shimachi et al. [195] developed a frame with a new axle-force-free (AFF) joint that can receive power from a tendon mechanism and then reproduce the haptic feedback to the user. Akinbiyi et al. [196] presented an intuitive augmented reality system through sensory substitution to provide force feedback enhancing the ability of the da Vinci system. Bethea et al. [23] developed a haptic feedback system in the form of sensory substitution for the da Vinci to perform surgical knot tying. Haptic feedback is also used in keyhole neurosurgery. For example, the ROBOCAST system [197] is a multi-robot chain together with a novel targeting algorithm that was designed to assist the surgical probe insertion during operations. This system consists of a multiple kinematic chain of three robots with a total of 13 DOFs, performing probe insertion to the desired target and following a planned trajectory. It also provides high targeting accuracy by combining haptic feedback inside the robotic architecture with an external optical tracking device.

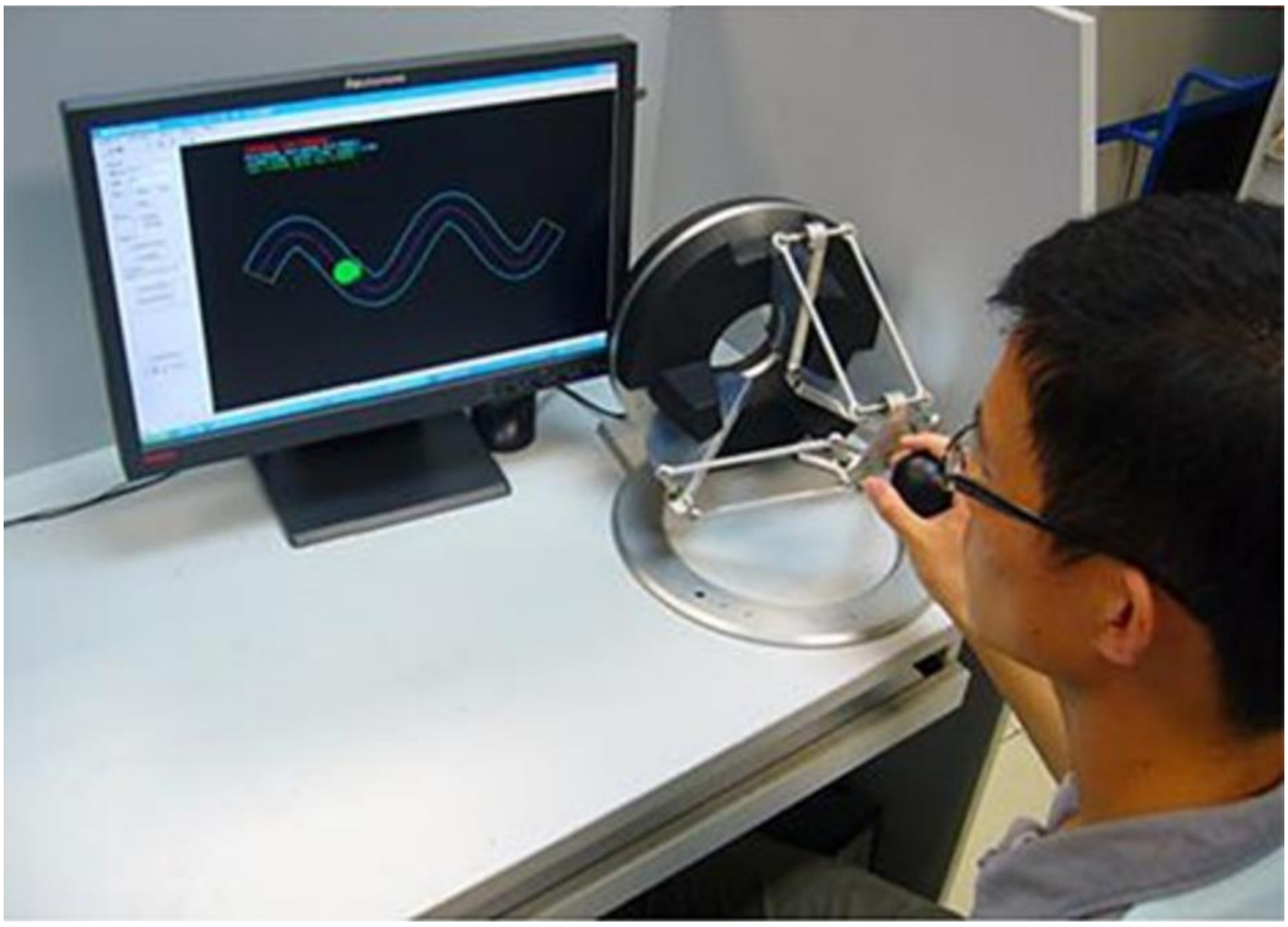

Figure 18. OMEGA haptic device. Reproduced with permission.[194] Copyright 2012, John Wiley and Sons.

In the effort toward the implementation of haptic feedback in surgical robots, it is worthwhile to highlight some other noticeable attempts: (1) neuroArm system [198] that provides a haptic corridor in the removal of glioma; (2) MAKO Tactile Guidance System [199] that is implemented in unicompartmental knee arthroplasty; (3) RCM robot [200] that provides a haptic sensation to the back of the surgeon's hand; (4) or a master-slave robot [201] with attached torque sensors that can provide force feedback in laparoscopic surgeries.

### 3.6. Advanced wearable eyeglasses for real-time vision and training

Wearable technology refers to any supporting devices that can be worn on the body or incorporated into clothes [202]. Wearable devices, especially wearable eyeglasses, have been rapidly developed to supply the increasing demand for general consumers in sport and entertainment sectors as well as in healthcare and medical applications [203]. Real-time vision is one of the most important requirements for the development of these devices, empowering real-time activities with recording capabilities [204]. The wearable eyeglass devices are also known as head-mounted displays (HMDs) that can be divided into two major categories (Virtual reality (VR) and augmented reality (AR)). Figure 19 overviews several commercial head-mounted displays that can potentially be used in surgical robotic systems.

VR HMD is a class of devices that can provide images via computer-generated imagery (CGI). Inside an HMD, a monitor typically LCD or OLED will display virtual images to the user's eye. A slight offset can be applied to create 3D imaging. The entire field of view of the wearer will be covered by the wearable device regardless of the visual direction from the eyes [205]. This type of HMD has a wide range of applications that include entertainment, sports, training, and manufacturing [206]. Many leading manufacturers have integrated VR technologies for their product development. The Oculus Rift (Oculus VR, Menlo Park, CA, USA) is a collection of VR headsets that was first released in 2016 for VR simulations and video games. The Samsung Gear VR (Samsung and Oculus VR) is also a VR headset that plays the role of the HMD and processors for use with Samsung Galaxy devices. The HTC Vive (HTC and Valve Corp.) can provide room-scale tracking and two wireless handheld controllers to interact with the environment and was released in 2016. The PlayStation VR (Sony Corp., San Mateo, CA, USA) which was released in 2016 offers VR solutions for gaming consoles and dedicated to PlayStation 4.

The second technology for wearable eyeglass is AR or mixed reality (MR) HMD (or optical see-through HMD). This technology refers to a kind of device that allows a CGI to be superimposed over the real-world view [207]. The front glass feature of this device is made of partly silvered mirrors known as an optical mixer that allows the wearer to look through it for the natural view and also reflect the virtual images. Google and Microsoft are the two most popular manufacturers in this area. Google Glass (Google, Mountain View, CA, USA) is a smart glass that has been available in the global market since 2014. It consists of a normal pair of eyeglasses and is equipped with a central processing unit embedded inside a frame, an integral 5 megapixel/720p video camera, and a prism display located above the right eye. This optical HMD can be wirelessly connected with the Internet and allows multiple communication modes such as natural language voice commands, built-in touchpad, blinking, and head movement [203]. Google Glass offers lightweight device configuration, user-friendly interface, and potential for hands-free control which may benefit surgeons in the operation room. However, several limitations should be addressed including privacy concerns and limited battery life. Microsoft HoloLens (Microsoft Corp., Redmond, Washington, USA) is a pair of mixed reality smart glasses that were first released in 2016. It is an optical HMD device with an adjustable headband for easy wearing. The HoloLens is equipped with powerful processors including a Holographic Processing Unit (HPU), a central processing unit (CPU), and a graphics processing unit (GPU), also an inertial measurement unit (IMU), a 2.4-megapixel video camera, a binaural audio system, projection lenses, and an internal rechargeable battery [208]. With these ultimate features, the HoloLens is an ideal smart glasses for a wide range of activities from gaming, virtual tourism, 3D development applications to interactive digital education and training including human anatomy and surgical applications.

Besides gaming and video, the HMD contributes to the engineering field by offering the virtual interaction between engineers and their designs. Furthermore, the maintenance of complex systems is becoming less troublesome thanks to the AR supplied by the HMD. In the field of training and simulation, the virtual feature of the HMD efficiently liberates trainees from the dangerous or expensive real-life situations such as flight simulation, military training, and surgical procedure exercises. Moving forward, several hurdles need to be overcome which includes cybersickness symptoms , expense of the device and accessories, and the limited availability of contents because of high development costs [209].

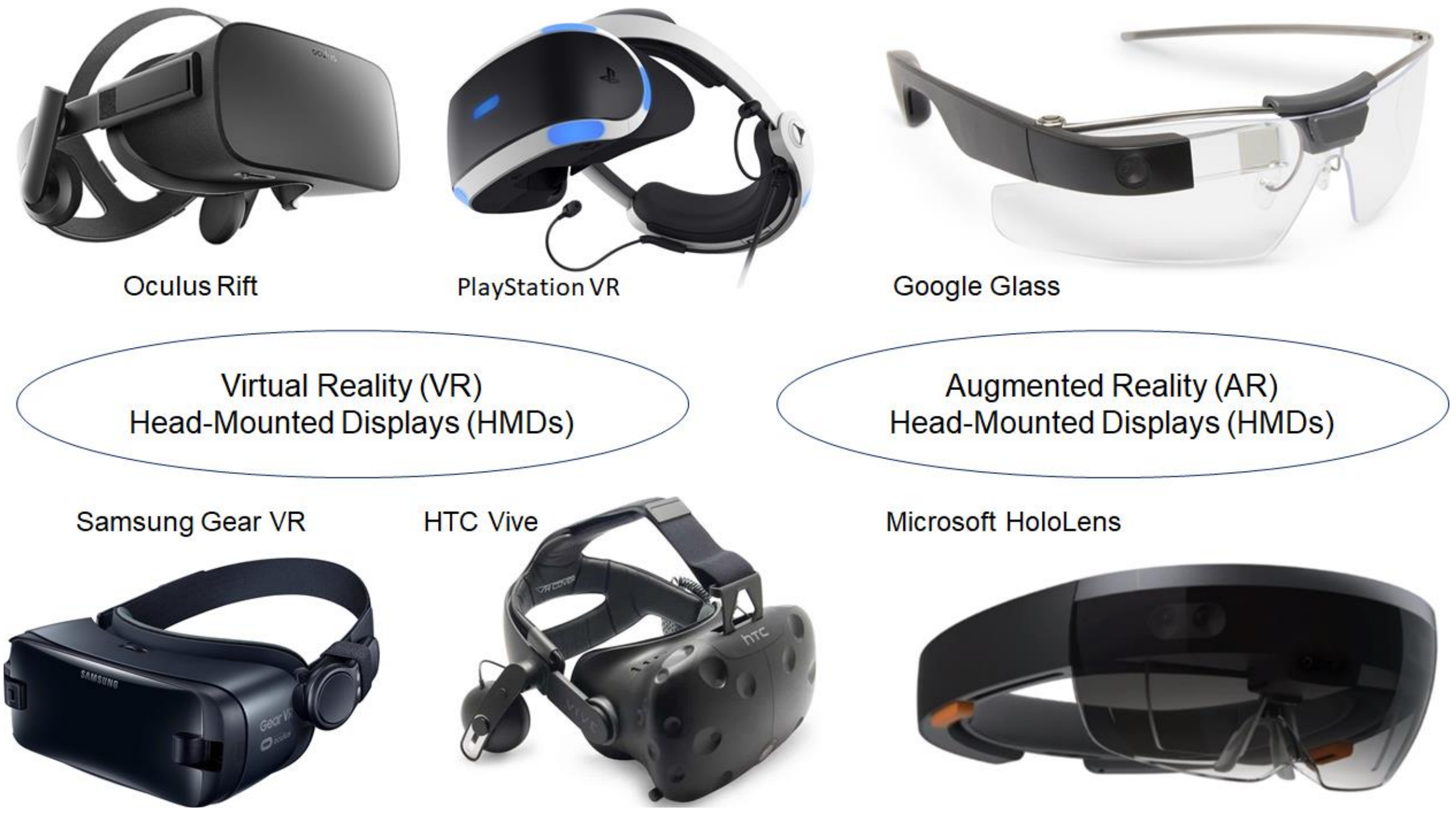

Figure 19. Several commercially available HMDs (Oculus, Facebook LLC., USA; Google LLC, USA; Microsoft, USA; Samsung, Korea; HTC Vive, HTC, Taiwan)

### 3.7. Advanced thermal sensors for surgical robots

The rapid growth of surgical robots requires concomitant development of associated sensory devices and techniques. Particularly, catheter thermal ablation is a common treatment for arrhythmias and other cardiac disorders [210]. Real-time thermal distribution feedback along the lesion is a critical factor during ablation procedures. The thermal feedback during the operation such as the record of electrophysiological data of live organs plays an important role in both surgical study and practice. Many research groups have concentrated on this promising field to benefit society. For example, Koh et al. [211] proposed an ultrathin injectable thermal sensor for cardiac ablation monitoring. The sensor comprises three golden micro-thermistors encapsulated in between a thin substrate of polyethylene terephthalate (PET) and a layer of photocurable epoxy (SU-8). These miniature sensors lessensed tissue damage during insertion and operation. Their sensing mechanism is based on the change of temperature coefficient of gold resistance where the temperature information is extracted from the measured resistance of each thermistor. This device is capable of real-time monitoring of temperature and thermal distribution at multiple depths in the myocardium during cardiac ablation procedures [212].

Several researchers have reported thermal sensors that can be integrated into a thin and flexible elastomer platform and then attached to the surface of target tissues or organs. Multifunctional sensors can be used to obtain data on thermal distribution, blood flow, contact pressure, and other physiological parameters [213-214]. However, the logged data only illustrates the surface situation of the target, being a major drawback of this approach. In clinical applications, Xu et al. [215] introduced 3D multifunctional integumentary membranes (3D-MIMs) for spatiotemporal cardiac measurements. This thin elastic membrane is fabricated from casting a layer of silicone elastomer on a heart model. The membrane is flexible enough to entirely embrace the epicardium of the heart while maintaining the normal cardiac function. The 3D-MIMs create a robust platform for multifunctional sensors to facilitate dynamic measurements and stimulation. In the case of thermal sensors, an array of golden serpentine thermistors is integrated into the stretchable structure of a 3D-MIM to measure the spatial temperature distribution of the heart surface. Kim et al. [216] presented a stretchable sensor and actuator webs for cardiac diagnosis and therapy. The web device is embedded in a thin bioresorbable layer of silk than completely envelop curved surfaces of the epicardium of the beating heart without any mechanical fixtures. The integrated sensors in the web provide real-time data on the strain, electrophysiological signals, and thermal mapping of large surfaces of the heart. In another approach, researchers in [217] introduced a multifunctional inflatable balloon catheter that includes a series of sensors, actuators, and semiconductor devices embedded into a commercially available balloon catheter. The stretchable and interconnected structure of components allows them to be peripherally wrapped around the balloon in both deflated and inflation configuration. This network mesh of all sensors provides real-time cardiac electrophysiological, tactile and temperature data.

There is an upward trend of using fiber optic sensors for thermal sensing and monitoring in surgical applications [218]. The FBG sensor is a class of optoelectronic devices that can provide a continuous pattern of the temperature distribution along its optical cable. The use of the FBG sensor is straightforward where the change of temperature will cause wavelength shifts, and it is usually configured in the form of an array to capture the spatial signal. Simple construction and long-term reliability are the strengths of the FBG array, but it also brings the coarse spatial resolution [219]. To overcome the inherent limitation of standard FBG sensors, chirped FBG (CFBG) sensors were introduced to enhance the capability of normal FBGs in terms of capture of the temperature pattern along the grating area with sub-millimeter spatial resolution [219-220]. The Bragg wavelength of the CFBG is not constant compared to a normal FBG but its variation in space offers wide bandwidth and better measurement of thermal distribution. This technology opens a promising field that incorporates fiber optic sensors into diagnostic probes or surgical tools to acquire more useful dynamic data of the treatment subject. Thermal sensors in combination with phase change materials have been also applied to variable stiffness structure for surgical robots [221-222].

### 3.8. Advanced pressure system for surgical robots

Maintaining working space for surgical instruments and visualization is a critical requirement during surgical procedures, especially for the endoscopic approach [223]. The development of insufflation systems remains the foundation of further progression on surgical robotics. Many manufacturers and researchers have introduced advanced pressure systems including insufflation function into surgical robotic procedures. For example, high flow insufflation units (Olympus Medical System Corp., Tokyo, Japan) is an intra-abdominal insufflation system that provides high-speed insufflation as well as automatic smoke evacuation. Its latest, the UHI-4, can produce a maximum flow rate of 45 L/min, has adjustable levels of smoke evacuation, and is equipped with a small cavity mode for endoscopic vessel harvesting [224]. The Olympus insufflation unit has been widely implemented in operating rooms and also being employ as an essential component to develop other advanced systems. Kato et al. [225] introduced steady pressure automatically controlled endoscopy (SPACE) using an insufflation system that can supply constant pressure of carbon dioxide ($CO_2$) during surgical procedures. This system is equipped with an off-the-shelf overtube, a standard flexible gastrointestinal endoscope, a commercially available surgical insufflator (UHI-3, Olympus), and a leak-proof valve. The SPACE overcame safety checks and has been used in esophageal submucosal dissection (ESD) on humans where a steady pressure inside the gastrointestinal tract was successfully maintained. The research team also reported the feasible usage of the SPACE in conjunction with a commercial surgical automatic smoke evacuator [226]. Both systems can simultaneously operate to reduce smoke while preventing the collapse of the targeted cavity.

AirSeal® System (CONMED Corp., Milford, CT, USA) is an intelligent and integrated access insufflation system for laparoscopic and robotic surgery. AirSeal® iFS consists of a valveless trocar that can produce stable pneumoperitoneum, constant smoke evacuation, and high flow insufflation [227-228]. The system is capable of maintaining low-pressure of the abdominal cavity in various laparoscopic procedures. Sroussi et al. [229] reported that gynecological laparoscopy is feasible at 7 mmHg pneumoperitoneum with the AirSeal® System. Such low pressure is analogous with standard insufflation (15 mmHg) and offers many benefits of a smaller amount of $CO_2$ that is absorbed by the body. La Falce et al. [230] also reported the surgical feasibility of the AirSeal® System at lower $CO_2$ pneumoperitoneum pressure (8 mmHg) in robot-assisted radical prostatectomy (RARP). Most significant hemodynamic and respiratory can be operated within safety limits during the whole surgical procedure. Another benefit of the AirSeal® System has been reported in its creation of pneumorectum [231] where the system can maintain a stable working space inside the rectum for transanal minimally invasive surgery (TAMIS).

PneumoClear (Stryker Corp., Kalamazoo, MI, USA) is a smart, multifunctional insufflation system that can provide insufflation with heated and humidified $CO_2$ and smoke evacuation [232]. This insufflator is designed to supply the maximum flow rate of 50 L/min and retain a stable surgical space while eliminating smoke for a clear intraoperative view and protection of operating room staff. The benefit of warmed and humidified $CO_2$ has been proven to noticeably reduce postoperative pain [233] and significantly mitigate hypothermia, reduce peritoneal injury, and decreased intra-abdominal adhesions [234].

There are a number of other noteworthy insufflation systems that can be used in surgical procedures. They include the ENDOFLATOR® 50 (KARL STORZ SE & Co. KG, Tuttlingen, Germany) [235] which is a high-performance insufflator with the integrated heating element; the EndoSTRATUS™ $CO_2$ Insufflator (MEDIVATORS Inc., Minneapolis, MN, USA) which is a versatile $CO_2$ pumping system that connects to both wall and tank sources; the NEBULAE™ I System (Northgate Technologies Inc., Elgin, IL, USA) that has a high flow laparoscopic insufflator system and can provide temperature-controlled $CO_2$ gas [236]; the GS2000 Insufflation System (CONMED Corp., Milford, CT, USA) can produce 50 L/min flow rate and provide continuous and constant pressure of body temperature $CO_2$ gas for laparoscopic surgery [237]; the $CO_2$EFFICIENT® endoscopic insufflator (US Endoscopy, Mentor, OH, USA) is a smart insufflator that can save $CO_2$ gas with the flow management system and prevent over-pressurization [238].

### 3.9. Advanced electrosurgery system for surgical robots

Electrosurgical technology has become an essential component for all specialized surgeries ranging from conventional surgery, laparoscopic surgery, and endoscopic surgery. Its working principle is based on high frequency (100 kHz to a few hundred MHz) and alternative electric current at various voltages (0.2 kV to 10 kV) which generates heat to cut, coagulate, dissect, fulgurate, ablate and shrink tissue [239]. The intensity of these thermal effects determines tissue behavior such as devitalization starting at around 60°C and vaporization of the tissue fluid at about 100°C. The main difference between electrocautery and electrosurgery is the intrinsic source of heat. The heating of electrocautery is exogenous, meaning that surgical effects are caused by a heated metallic instrument compared with the endogenous source of electrosurgery generated by the current flow in the tissue itself [240].

There are three core techniques of electrosurgery: diathermy, monopolar, and bipolar. Diathermy is a therapeutic method that applies a high-frequency electric current to the body and then generates heat to the targeted tissue. The use of heat offers many advantages over the conventional knives as it can relieve pain, increase blood flow, accelerate healing, minimize inflammation and reduce fluid retention [241]. Diathermy can be divided into three major categories: shortwave, microwave, and ultrasound [239]. The shortwave diathermy technique uses high-frequency electromagnetic energy with a frequency generally of 27.12 MHz and a wavelength of around 11 meters to generate heat. This approach is typically prescribed for pain relief and muscle spasms. To perform therapy, the treatment area is placed between two electrodes from a shortwave device. Microwave diathermy uses microwaves with a frequency above 300 MHz and a wavelength of less than one meter to heat up the tissues without skin damage. This therapeutic treatment is suitable for superficial areas because of the poor depth of penetration of microwaves. Ultrasound diathermy exploits high-frequency acoustic vibrations to generate heat for deep tissue treatment. Each type of muscles or tissues has a distinguished sensitivity with a certain ultrasound frequency, facilitating the treatment for selected musculatures and structures. This therapy is usually recommended for muscle spasms, lithotripsy, and hemostasis.

The monopolar technique consists of an active electrode at the instrument tip and a neutral electrode attached to the body's skin (Figure 20). The heating effect happens at the small contact area between the active electrode and tissue although there is an acceptable amount of heat on the large surface of the neutral electrode [242]. This electrosurgical mode is widely used in many surgical procedures due to its simple approach and great handling properties. However, the current flow through the body may create negative effects on pacemakers or metal prostheses.

The bipolar technique is typically equipped with two electrodes integrated at the instrument tip (Figure 20). The heating effect mainly occurs in the confined tissue area between the two electrodes [242]. Compared to the monopolar approach, this technique offers better safety because of the narrow area of the current flow. However, the bipolar approach is limited in several surgical applications.

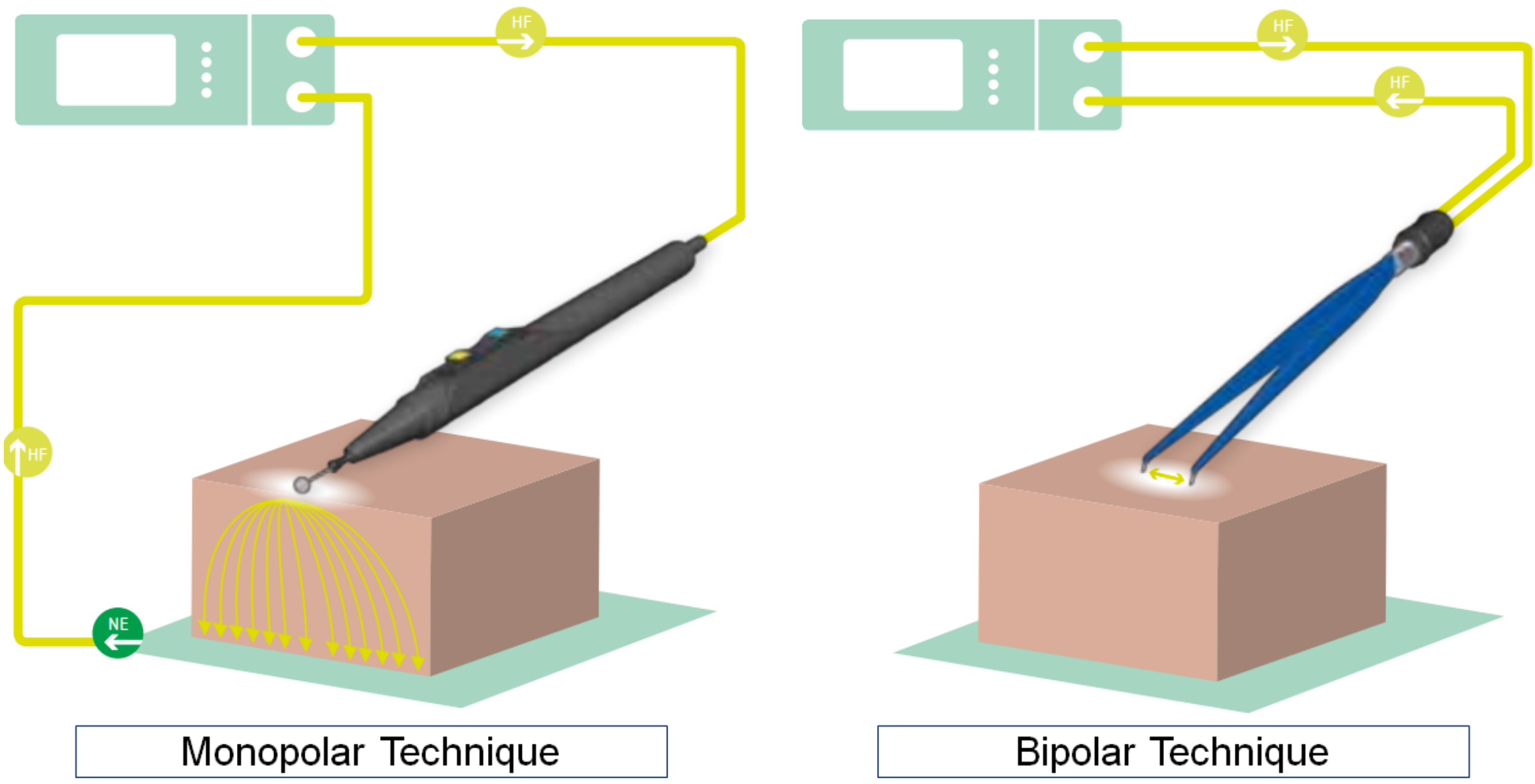

Figure 20. Principle of monopolar and bipolar techniques [240] (HF: high-frequency current flow, NE: neutral electrode)

Energy-based instruments are increasingly used in laparoscopic surgery because of better hemostasis and less lateral spread of heat. The HARMONIC ACE®+7 (Ethicon Inc., Cincinnati, OH, USA) is an ultrasonic coagulating shear that provides the ability to seal vessels with a diameter of up to 7 mm [243]. Incorporated with Adaptive Tissue Technology, this device can intelligently adjust the supplied energy to create better surgical precision and to reduce the risk of thermal injury. LigaSureTM (Valleylab Inc., Boulder, CO, USA) is an electrothermal bipolar vessel sealer, using a combination of pressure and energy to create vessel fusion and is capable of sealing vessels with a diameter up to 7 mm [244]. The device automatically halts the energy when the sealing procedure is completed. ENSEAL® G2 (Ethicon Inc., Cincinnati, OH, USA) is a bipolar vessel sealing system that combines energy and compression to simultaneously seal and transect vessels up to 7 mm in diameter. The ENSEAL® G2 outperformed the LigaSureTM in terms of lower blood loss, less tissue sticking, and higher burst pressure in sealed vessels [245]. THUNDERBEATTM (Olympus Medical System Corp., Tokyo, Japan) is a hybrid advanced energy system that simultaneously provides ultrasonic and bipolar energy for dissection and hemostasis for vessels [246].

Relating to electrosurgical generators for surgery, various competitive systems are currently available in the market as they offer multiple electrosurgical modes and are equipped with associated surgical tools. They include: (1) The VIO® 300 D from Erbe Elektromedizin GmbH, Tubingen, Germany; (2) The Valleylab™ FX8 energy platform from Valleylab Inc., Boulder, CO, USA; (3) The ESG-100 from Olympus Medical System Corp., Tokyo, Japan; (4) The gi4000 from US Endoscopy, Mentor, OH, USA; (5) The System 5000™ (CONMED Corp., Milford, CT, USA); (6) The MEGADYNE™ MEGA POWER™ (from Ethicon Inc., Cincinnati, OH, USA); (7) The Bovie OR|PRO 300 (from Symmetry Surgical, Antioch, TN, USA).

### 3.10. Advanced control algorithm for surgical robots

Bilateral teleoperation is a surgical technology that enables the surgeon to perform an operation from a remote distance. This approach is based on two-way communication between the master console (commands) and the slave end-effector (surgical tools). Depend on the specific characteristics of each teleoperation system, there are a number of control algorithms that can enhance the system stability during the operation with low or no nonlinear hysteresis and backlash [247-250]. For example, a four-channel force-velocity architecture proposal provides no communication delay between the master and the slave side [251]. In the case of dealing with constant communication delay teleoperation systems, several architectures are recommended such as wave variables, proportional-integral-derivative (PID) controller, feedforward, and adaptive feedback algorithm. Typically, there are three approaches that are proposed for bilateral teleoperation systems with time-varying communication delay. They include the time domain passivity approach (TDPA), passive set-position modulation (PSPM), and hierarchical two-layer approach [247].

Interaction control is a control technology that has been heavily involved in the dynamic constraints of the environment and can be divided into active and passive interaction control. The passive interaction control relates to the position of a manipulator without force information while the active interaction control focuses on the dynamic displacement-force relation – also known as compliant control. The compliant control allows the robotic system to safely interact with unstructured environments via transferred energy to the environment rather than position or force [252]. Admittance control, on the other hand, is formulated by an inner position loop and an outer force loop, creating a robust but low accuracy algorithm. In contrast, impedance control has an opposite configuration to produce higher precision, but a high impedance is hard to achieve. The selection of control algorithms highly relies on the required robustness/accuracy of specific manipulators.

Recent decades witness the incredible development of flexible manipulators for robotic platforms to substitute heavy and bulky approach as from conventional robotic manipulators. Flexible robots offer many advantages for a lightweight, low cost, better transportability, and high-speed operation compared to their conventional counterparts. However, most flexible manipulators have an intrinsic problem of instability and vibration at the robotic tip that causes inaccuracy in trajectory and positioning control [253-256]. Inspired by the great potential of flexible manipulators, many researchers have proposed advanced control schemes to overcome these obstacles, improving robot stability and accuracy [257-260]. These control techniques can be categorized into two types based on the ways that the input data are generated. The first category is a model-based control strategy with a dominant type being a feedforward controller which requires no feedback information from sensors during the control implementation. However, the nonlinearities and uncertainties originated from system variations and the external environments are normally not taken into consideration. Several model-based controls have been proposed in the literature. They include: *(1) Feed-forward control* which is a simple control technique without using any feedback information. This scheme is recommended for repetitive tasks as it is not able to deal with the variable payloads of flexible systems [261-264]. *(2) Boundary control* which is suitable for trajectory tracking tasks where the value range of model parameters are well defined. This control approach can deal with variable payloads or non-linear dynamic systems [253, 265-266]. *(3) Optimal trajectory planning* is a control strategy that is mainly used to find optimal paths while maintaining minimum vibration for the flexible system. The performance of this technique heavily depends on feedback control [267]. *(4) Input shaping technique* intercedes the input signals in real-time to cancel unexpected vibration, providing robust stability of the system. In return, it requires an accurate mathematical model which is difficult to achieve in many flexible systems [268]. *(5) Predictive control* generates input signals based on the prediction of future output signals [269]. This technique provides robust performance even in the presence of delay feedbacks or non-linear systems. Nonetheless, the predictive controller is normally slow and unstable, leading to the development of a general incremental predictive controller (GIPC). The GIPC significantly improves stability, providing fast response and robust performance [270].

The second category is non-model-based controllers where input signals are regulated by real-time feedback from sensors to minimize the vibration and system delay. *(1) Position feedback control* utilizes displacement information from sensors to manipulate the flexible systems. This approach can adapt to the changes of system states and effectively eliminate the vibration mode without destabilizing target mode. However, it has trouble dealing with time-varying frequency components [271-273]. *(2) Linear velocity feedback control* shows effective damping as well as maintaining the stability of the closed-loop system in a trade-off of high control effort over the operating frequencies [274]. *(3) PID approach* is the simplest control method that has been used in many mechanical and robotic systems due to its simple structure and reliability. This controller is not able to deal with nonlinearities and disturbances. It is often combined with other techniques to improve system performance [275]. *(4) Repetitive control* offers a simple approach that can quickly remove vibrations by tracking zero steady-state error periodic references. However, this technique is not able to adapt to system changes [276]. *(5) Fuzzy logic control (FLC)* is another technique that can be implemented into any systems that do not require an accurate mathematical model. Although it has an easy design and implementation, there exist difficulties to tune the control parameters that sometimes lead to instability in the system [277-278]. *(6) Adaptive control* is a real-time feedback control that can online-tune the model parameters to adapt to the change of environment, uncertainties, and disturbances. One of the main challenges for this scheme is online feedback during the compensation that is hard to achieve, especially in surgical robotic systems where sterilization and miniature size of flexible tools are necessary [279-280]. *(7) Neural Networks (NN) based control* is a technique used when dealing with unknown system dynamics. Users can eliminate the use of complex mathematical models from the system. However, this control scheme requires a sufficient amount of training data set [281].

## 4. Automation for Robotic Surgery

Automation for robotic surgery involves the use of a robot to perform surgical tasks under the control of partial or no human guidance. Market leader Intuitive Surgical reported that surgical automation had a growth rate of nearly 20% between 2017 and 2020 [282]. Nowadays, autonomous surgical systems have shown a feasibility to replace several surgical tasks from the surgeon while maintaining high precision and efficiency during surgical procedures. The need for automation in surgical robotic systems orginates from the fact that surgeons may be overworked and fatigued resulting in human error which may be harmful to patients [47, 283]. Advanced technologies in tele-operation now allow the surgeon to carry out remote control of surgical procedures away from dangerous radiation from X-ray fluoroscopy [284]. Recent novel endoscopic robots or robotic catheter systems have been equipped with intelligent control algorithms to enable an automatic learning and assistance of multifaceted mappings from the proximal handle to the distal tip of the surgical tools, avoiding complex professional training that is associated with time and cost [285]. In addition, robotic devices now can automatically eliminate the tremors of the surgeon's hands, enabling a precise control of the surgical tools compared to traditional surgery [28, 286-287]. Automation in surgery through wristed manipulators is believed to provide consistent quality across surgical cases with greater dexterity in surgical tools than a human-controlled tool.

Automation in robotic surgery can be categorized into different levels, depending on the specific surgical procedures. It can be no autonomy, robotic assistance, specific task autonomy, conditional autonomy, high autonomy, and full autonomy [48, 288-289]. The development of autonomous surgical systems requires many strict constraints to meet safety concerns and to maintain efficacy of surgical procedures. Most existing surgical robots are slave machines where intelligent predictive analytics to supervise an entire procedure are lacking, thereby preventing them from working independently. With the no autonomy level, teleoperated robots or prosthetic devices follow the surgeon’s command through a user control interface [51]. This level of autonomy is prevalent and has been shown to be a good method without the use of any cognitive decision making. With the robotic assistance level, the robot is controlled by the user via guided mechanisms. With the autonomous robot level, surgical tasks such as surgical suturing can be independently carried out using a preprogramed curve under the surgeon's supervision [290]. The intermediate level provides conditional autonomy where the surgical robot can intelligently generate and execute preoperative plans. For this type of system, the surgeon is able to control generated plans prior to its execution. Therefore, the robotic system can safely complete the surgical task. With the high autonomy level, the surgical robot can make decisions under the supervision of a qualified doctor during the procedure. A full-autonomy robot has currently not been described in the literature although this idea has been proposed in science fiction movies where no human is involved in the operation [291]. At this level, the surgical robot can completely perform a surgical task with safety and accuracy without the assistance of a surgeon. A key requirement in the development of automatic surgical robotics is the capability of replicating the surgeon's sensorimotor skills. With advanced development in medical imaging, sensing techniques such as position and force sensing, and actuation; it may be possible to develop a surgical robot towards full autonomy [283]. However, ethical barriers and legal concerns need to be addressed before fully intelligent robots can be safely introduced.

## 5. The advantage and disadvantage of automation in robotic surgery

High-level automation employed in surgical robotic systems may offer higher accuracy and speed during surgical interventions, especially with intricate procedures which require fine dissection for sparing nerves and vessels [282]. For example, the ARTAS hair restoration system (Restoration Robotics Inc., San Jose, CA, USA) (Figure 21) can identify and remove folliculitis from the head in a fully autonomous manner under the support of the real-time image guidance system. This robot offers higher accuracy and safety compared with manual operation by surgeons [292]. Benefits of surgical robot automation in biopsy or therapy delivery cases has been demonstrated by the increase of consistency and dexterity during surgical treatments. With robotic automation, surgical tools can be navigated at an optimal angle to reach the tissue target while avoiding unexpected collisions with surrounding organs as well as reducing organ damage. One notable autonomous surgical interventional system is the Hansen Medical Sensei System. It is automatically controlled by intelligent system where motion of the catheter tip is pre-programmed to automatically adapt with the environment, allowing the clinician to achieve a faster time to reach the target [293]. The da Vinci Xi generation (see Figure 22) is the first surgical system that can feature automated docking, instrument positioning, and camera adjustments. It can automatically follow the adjustments made with the robot while maintaining the position of the patient relative to the robot [45]. Intelligent systems are also integrated into surgical instruments to quickly respond to a change in the surrounding environment by using onboard smart sensors. During the operation, medical data stored in the system from the smart sensors can be used to provide up-to-date status of the instrument, enabling the surgeon to make appropriate decisions to solve problems [294]. Modern surgical systems with automation are capable of providing assistive tasks such as suction, tissue retraction, irrigation, and staple application that can substitute skilled surgical staff [295]. Machine learning, a form of AI, enables the capability of learning from prior experiences [296-297]. AI has been introduced to the surgical robotic system to perform trained algorithm tasks, predict outcomes, and to change directions in real-time based on previous experiences [298]. AI has an ability to automatically capture and analyze patient information to formulate an initial diagnosis and potentially suggest treatments [297].

Nowadays, full autonomy in complicated surgical applications remains extremely challenging due to limitations in technology and the question of ethics. Supervised autonomy, therefore, seems to be a suitable approach in most existing surgical procedures compared with full autonomy [300-301]. Human skills and experiences can contribute to decrease the adoption barrier for robotic systems and ensure successful execution with supervised autonomy. Although deep learning methods have achieved impressive results and outperformed human skills in some cases, the development of robotic system intelligence to a similar human level, is currently unlikely due to the complexity to build [302]. While surgeons are able to decompose and simplify the whole surgical procedure with many steps, most robotic systems can only perform a single task. Therefore, the use of a surgical robotic system with automation during operations can be controversial. Supervised autonomy allows human experts to make critical judgments to achieve safe and effective surgical outcomes [303]. One of the earliest examples for supervisory controlled robots is the TSolution One Surgical System (Figure 23). The robot is able to do autonomous bone drilling based on the surgical plan, derived from pre-operative CT images [45]. Another example is the AutoLap camera-handling robot (developed by MST Medical Surgery Technologies Ltd). This system provides feature-tracking with image guidance during laparoscopic procedures [45].

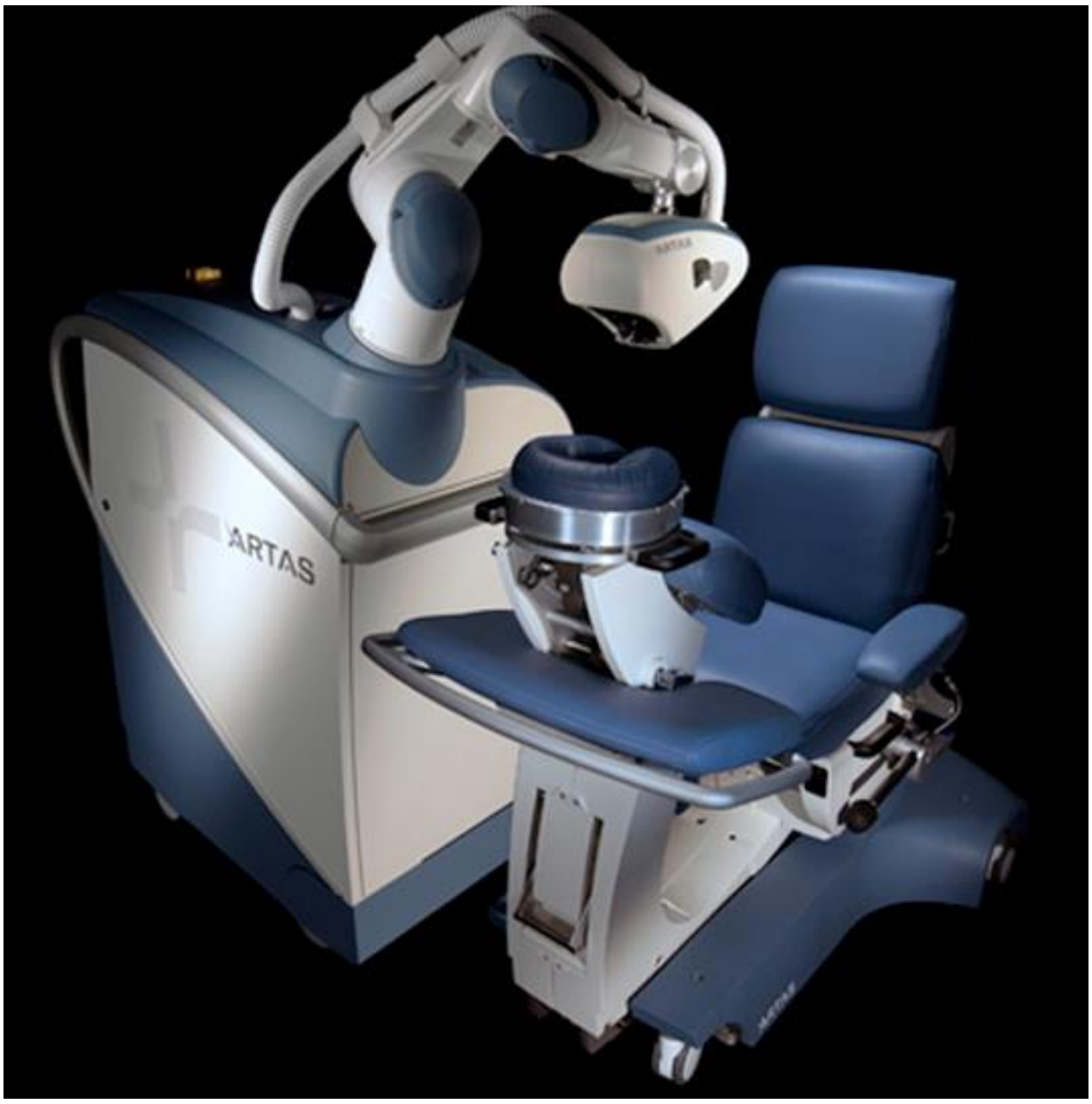

Figure 21. ARTAS hair restoration robot. Reproduced with permission.[299] Copyright 2015, Elsevier.

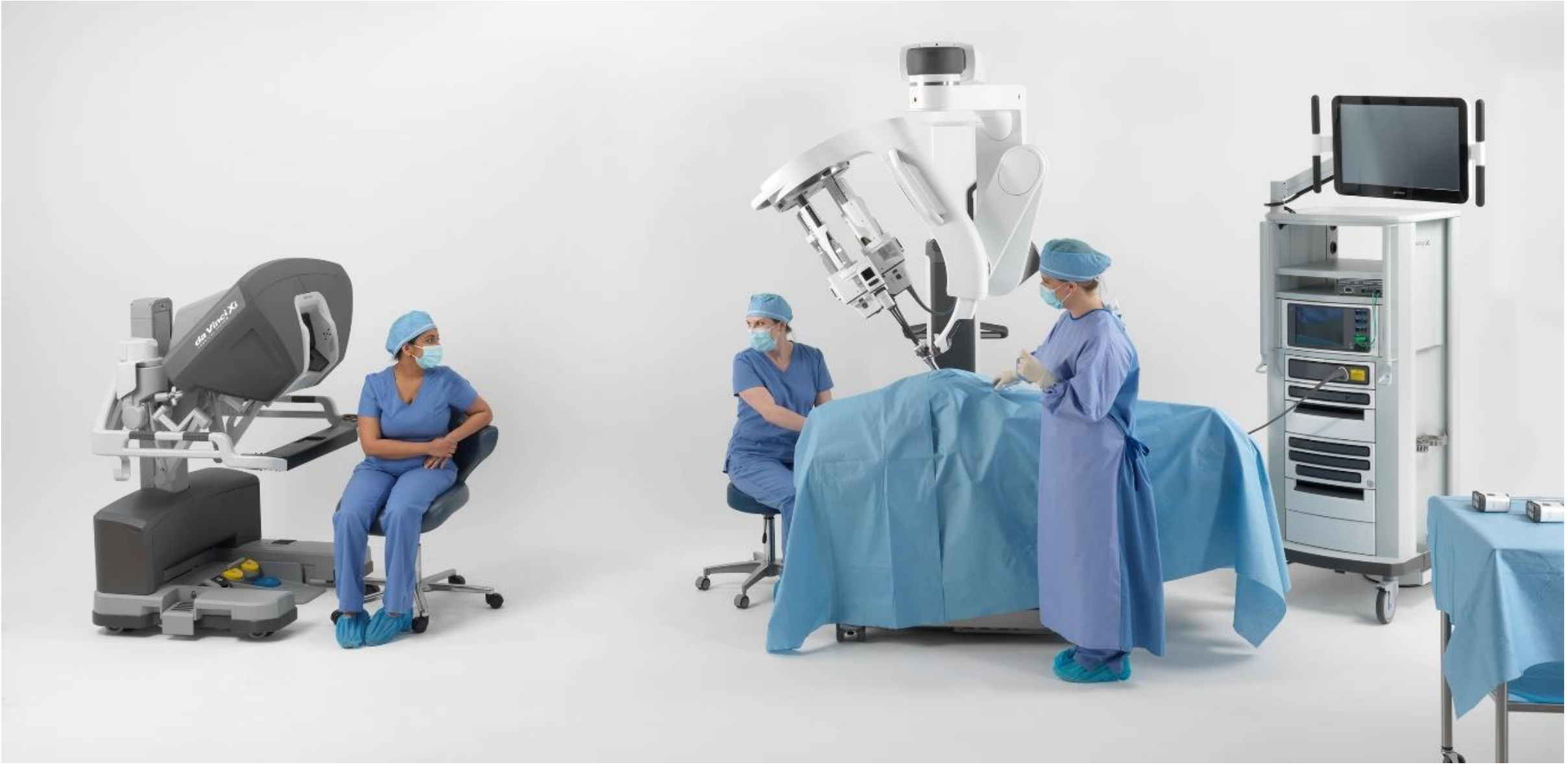

Figure 22. The da Vinci Xi teleoperated system with a self-adjusting operating table. ©[2019] Intuitive Surgical, Inc.

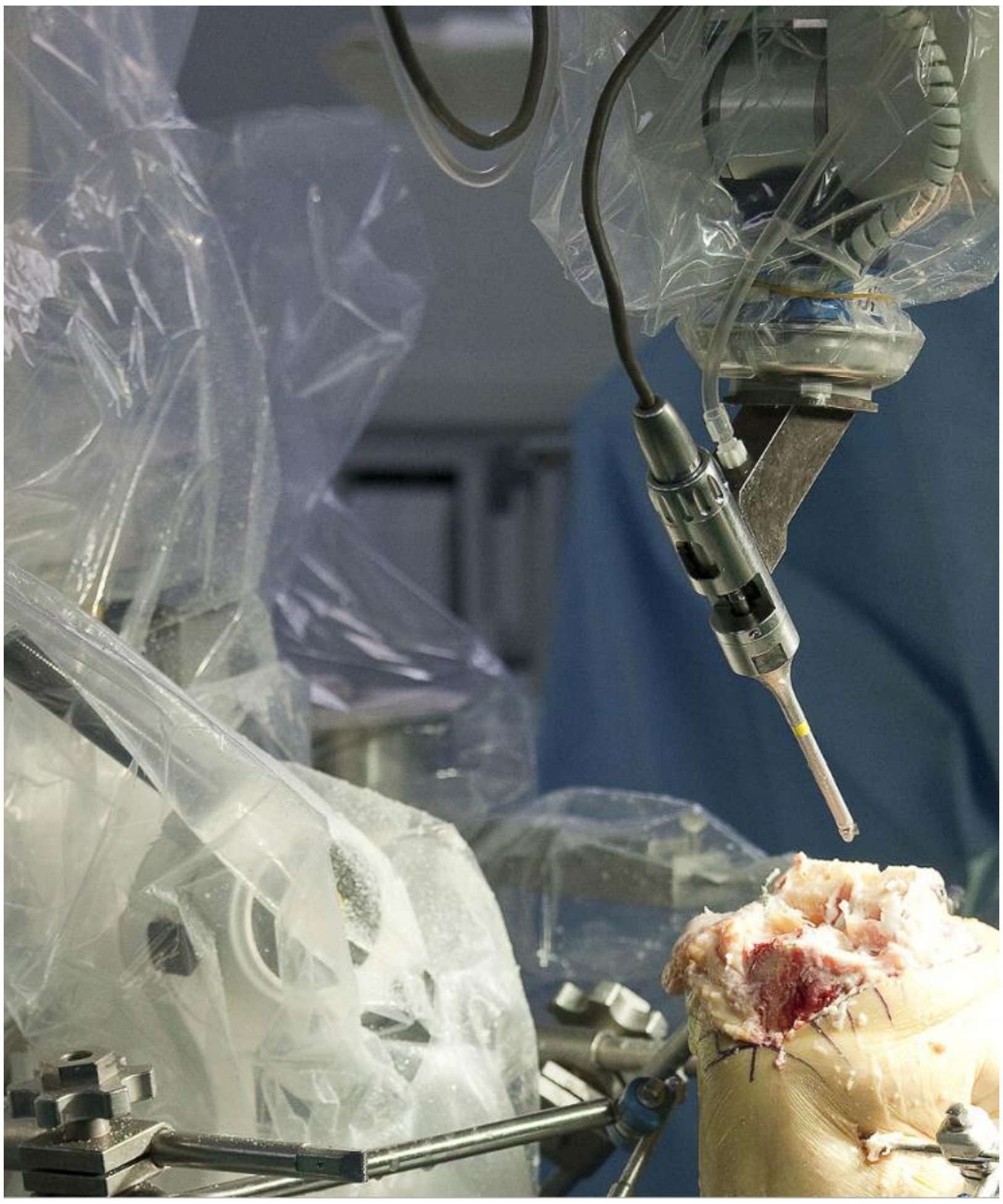

Figure 23. Think Surgical's TSolution One system (formerly called ROBODOC) offering image-guided (autonomous) knee arthroplasty. Reproduced with permission.[304] Copyright 2015, Elsevier.

Despite the merits, automation in surgery has potential major drawbacks which include potential failure in mechanical systems, electronics, and software. In addition, technological gaps to manage all problems associated with autonomous surgery are still unsolved. As a result, the fears of litigation discourage many surgical companies towards the development of new types of surgical robots. Patient safety should be the gold standard that has to taken into consideration if automation is involved during a surgical procedure [305]. Another disadvantage of automation is the high cost which may limit access for low-income patients or patients in rural and remote areas. Surgical robotic system with high level of autonomy requires regular maintenance and potentially expensive equipment upgrades [306]. The operative time may be longer compared to conventional systems, especially in the learning curve phase, partly related to the complex set up of the system. Despite limitations and constraints of technology, regulatory, ethical, and legal barriers; the development of advanced intelligent systems can achieve greater levels of automation in surgical systems in the future.

## 6. Potential surgeries with automation

Although surgical robots have been used in many hospitals and medical services around the world, most of them are currently teleoperated robot-assisted systems with no automation. Automation in robotic surgery can offer many benefits over the conventional manual approach such as higher accuracy and reliability, improved dexterity, and speed. Autonomous systems for surgery can be used to perform simple surgical tasks such as preparation, suturing, cutting, and puncturing. This section will describe a short overview of potential surgeries with robotic automation in different types.

Radiosurgery may be the ideal type of surgery to adopt full automation. The radiation beams can be precisely directed towards the targeted tumours while avoiding healthy tissues [307]. This approach does not require any physical interaction between the surgical devices and patient during the operation. A few commercial systems such the CyberKnife, the Novalis and the Gamma Knife have successfully used automation in surgery [308]. However, they require a pre-operative scan of the targeted areas to plan the direction of irradiation. With the CyberKnife system, the surgeon can define the treatment area using advanced imaging systems such as MRI or CT scan. An intelligent program plans the delivery of the optimal radiation dosage to the targeted tissue. Optical markers are attached on the patient’s body to track breathing motion while the tissue target positions are updated [309]. There is further development of this system to achieve full autonomy.

Orthopedic surgery and neurosurgery can be achieved with surgical automation. Advanced image guidance based on the Computer-Aided Design and Manufacturing (CAD/CAM) paradigm has been successfully utilized in clinical trials [310]. The CAD/CAM paradigm creates surgical plans and executable tasks in advance based on CT or MRI pre-operative image resulting in superior precision and lower rates of failure [311]. To achieve a higher level of automation, additional safety mechanisms are needed to protect the patient from system failure while improvement in control algorithms can contribute to achieving complete autonomous task execution. Orthopedics may be one of the best placed types of surgery equipped with an automation approach because the surgical tasks such as drilling into specific areas of the bones can be predefined prior to the implant surgery. For orthopedics and neurosurgery, the bones are non-deformable and stiff and therefore this method can be easily implemented with preoperative plans to achieve higher accuracy and high consistency compared to cases with soft deformation tissue.

Automation in cardiovascular surgery potentially offers many benefits to the surgeon. Current limitations of cardiac surgery are the ability to determine the target location accurately, automatically adapt to the surrounding environment, and precisely control the applied force to the tissue. For electrophysiology mapping and cardiac ablation, automatic robotic catheters can overcome difficult tasks during the operation such as the ability to follow the moving tissue target while maintaining desired interaction force [312]. Advances in actuation and imaging technologies have revolutionized the use of automation in cardiac surgery. This includes the use of magnetic actuation and haptic vision that wirelessly transmits power to the catheter tip via an external magnetic field while imaging systems can offer feedback to provide autonomous navigation and learning [313]. The catheters will automatically approach the heart through the femoral or brachial artery or vein under the use of X-ray visualization for navigation assistance. Researchers from Harvard Medical School recently developed positively thigmotactic algorithms that can achieve autonomous navigation inside the human heart with a low contact force with the soft tissue, following the tissue walls to reach the desired target. New sensing modalities (haptic vision) which combines machine learning and image processing algorithms to form hybrid imaging and touch sensing have also been implemented [313]. The haptic vision served as sensory input to achieve wall following while controlling the interaction force applied by the catheter to the tissue. This method was validated in autonomous navigation through in-vivo experiments and was also compared with manual navigation. As autonomation offers many advantages, this technology can be expanded to perform more difficult tasks.

Autonomation can potentially improve flexibility, accuracy, repeatability, and blood loss in cryoablation procedures for kidney tumors. Laparoscopic cryoablation is a minimally invasive technique to treat a small peripheral tumor in the kidney. During the procedure, a laparoscopic tool and camera are inserted into the human body via laparoscopic ports. An ultrasound probe is then introduced to scan the entire kidney so that the optimal area and location of the target can be identified [314]. To achieve a high success rate, good clinical skills and hand-eye coordination are important factors. A semi-autonomous model-independent needle insertion technique was introduced to improve the efficacy of this surgical procedure [315]. A novel vision-based semi-autonomous collaborative bimanual procedure demonstrated a reduction in the risk of kidney fracture with more accurate positioning of the tool while maintaining accurate tracking of the deformed tissue.

Another potential area of application for partially automated surgery is microsurgery of the eye. Laser photocoagulation is a common treatment for retinal disease as it applies patterns of multiple burns to seal the leaking blood vessels in macular edema or to impede the growth of abnormal blood vessels in retinopathy [316]. Robotic technology has been implemented to improve the accuracy to achieve optimal clinical outcomes and to reduce the operative surgery time. Recent automated approaches have resulted in a rapid delivery of multiple but shorter time pulses with predefined spots using a galvanometric scanner. A hybrid retinal tracking system has been described that incorporates a fully automated approach featuring an improvement in accuracy and possessing an ability of real-time compensation for eye mobility. By combining digital fungus imaging in real-time, the system can automatically perform computer-aided retinal photocoagulation. This technology is now commercialized as PASCAL and Navilas® [317]. Micron, a handheld micromanipulator, was also developed to perform laser probe automated scanning for retinal surgery [317]. This system features a new approach for automated intraocular laser surgery with an expanded range of motion for automated scanning while adding degrees of freedom to accommodate use via sclerotomy [318].

## 7. Essential technologies, challenges, and the future of surgical robots

The growth of the aging population and the increase of life expectancy means an increase of demand for surgery in the future. Robot-assisted surgeries promise to alleviate the surgeon shortage to perform advanced medical procedures. With the growing adoption of new technologies, it is possible that eventually any future surgical procedure will contain at least one robotic component in the operation. Recent success in the development of autonomous systems such as self-driving cars or manufacturing processes have increased the trust from communities for the future of autonomous surgical systems. Although the development of autonomous surgical systems has not been widely reported, early successful clinical trials with automation have demonstrated many potential benefits over existing surgical techniques [319]. However, the implementation of automation in surgical procedures is still challenging due to the technological gaps. There are three main stages for an automation-relevant future surgical procedure: (1) the understanding of surrounding environments through the use of tactile/force and position feedback with imaging techniques; (2) the establishment of a plan to adapt with changes of environment; (3) the execution of surgical tasks [320]. This is a repetitive loop to carry out surgical tasks until the process is completed. At each stage of an operation, specific technologies are required. It is believed that advanced technology in sensing and actuation will play a vital role towards the development of autonomous surgical systems [321].

Another potential technology for the development of a complete and intelligent autonomous surgical robot is AI which may provide better assistance and support for the surgeon as well as the ability to execute complex surgical tasks in a safer and more efficient way. Along with the development of tele-heath systems and online data monitoring, AI will be able to solve problems faster than human doctors, suggest better solutions, and enhance communication between patients and clinicians [322]. In addition, autonomous surgical robotic systems will be equipped with advanced control algorithms, deep learning ability, and reinforcement of learning. This can lead to suggestion of optimal insertion points and trajectories or planning of insertion points to the surgeon or providing the ability to react with surgical events in an appropriate way [323]. Cognitive surgical robots, which are intelligent systems, can manage and control the workflow of surgical procedures as well as predict the mental state of human collaborators (surgeons and assistants) to provide better support during the surgery. Supervised autonomy, on the other hand, requires advanced technologies such as algorithmic clinical support, haptic devices, VR/AR, 3D vision and other assistive tools to support the surgeon during surgery. This technology may be the most prevalent level of autonomy utilising in the near future. Researchers at Boston Children's Hospital and Harvard Medical School demonstrated an intelligent platform that can automatically navigate into the human heart valve with better performance compared with previous approaches [324].

Safety is a paramount concern for most robot-assisted surgical procedures that needs to be addressed in order to achieve wider acceptance. Although surgical robots have been developed and successfully validated by many clinical trials, fears and uncertainties with the environment and tools when working with robots in an autonomous mode still exists. This may be related to the fact that the healthy tissue or organs are at potential risk of danger if the surgical instrument is in close proximity. Although preoperative planning would be a positive solution, the change of environment is uncertain and therefore real-time accommodation to adapt with these changes would be an important addition for the autonomous approach [325]. Surgical robots must be able to understand the surrounding environment in order to adapt to the dynamic changes of the surroundings and adjust mission execution instantaneously via real-time feedback using advanced tactile/position sensing technologies.

Autonomous surgical systems are expected to have functions for error detection while dealing with critical situations. The cost to maintain and upgrade equipment in future autonomous surgical systems would be high, especially given regulatory compliance, and therefore it may be a barrier that need to be taken into consideration. New smart user interfaces to manage, observe, and control autonomous surgical systems with high efficiency and robustness are also vital. The Monarch system (developed by Auris Health, Inc., USA) to perform lung cancer diagnosis and treatment under fluoroscopy is a typical example of future autonomous surgical system [326].

The U.S. FDA reviews and approves robotic-assisted devices through the 510(k) process. Any new surgical robotic systems are classified as high-risk devices which require the most rigorous PMA (premarket approval) governing pathway. A PMA device takes an average cost of $94 million and 54 months from first communication to market implementation [327]. At higher levels of autonomy, autonomous surgical systems would need a certificate that is equivalent to a human surgeon. Therefore, approval may be a significant barrier to implement automation in future surgical systems. Increasing the level of autonomy for surgical procedures may also raise questions about the responsibility if errors or failures occur. Future autonomous surgical systems should also be covered by a risk management process with regulatory, ethical, and legal frameworks, in order to achieve desired surgical outcomes.

Nowadays, autonomous machines such as self-driving cars and drones have become popular in daily life, with many reports of their successful implementations. The acceptance of risk from autonomous robots for medical applications is expected to increase in the near future. Many research groups around the world are collaborating towards the development of surgical autonomy. The high-tech sensors and intelligent algorithms should be designed to replicate the sensorimotor skills of expert surgeons in both sensing precision and resolution. In addition, novel imaging modalities should be available to go beyond human ability in complicated cases. On the other hand, emerging international safety standards are needed to support the commercialization of autonomous surgical robots. Until these standards are widely accepted, supervised autonomy seems to be the most promising level of autonomy to avoid regulatory hurdles. In the pursuit of autonomous systems in surgery, alternative methods for new surgical approaches which enhance surgical accuracy and reliability must be considered. Although further studies are required, initial data have revealed successful use of automation in robotic surgery can provide better results compared with conventional procedures.

**Acknowledgements**

The authors acknowledge support from the UNSW Start-Up Grant (PS51378) and UNSW Scientia Fellowship Grant (PS46197).

**Table 1. Available surgical robotic platforms in practice**

| **Platform** | **Developer** | **Specialization** | **Features** | **Approval status** |
|---|---|---|---|---|
| **Commercial surgical robotic systems** | | | | |
| da Vinci® Surgical Systems [329] | Intuitive Surgical Inc., Sunnyvale, CA, USA | Laparoscopic, thoracoscopic, prostatectomy, urology, gynecology, cardiotomy, pediatric, revascularization, transoral otolaryngology, cholecystectomy, and hysterectomy | Master-slave tele-manipulation; 4 robotic arms: 3 surgical tools (7 DOFs each) and 1 3D-HD camera; wide range of fully wristed instruments | FDA in 2000<br>CE Mark in 2017 |
| Flex® Robotic System 106] | Medrobotics Corp., Raynham, MA, USA | Transoral surgery of the oropharynx, hypopharynx, and larynx | Flexible robotic scope; the inner mechanism follows the outer mechanism; 2 flexible instruments and 1 HD camera | FDA in 2015<br>CE Mark in 2014 |
| TSolution | THINK Surgical | Total hip | One milling | FDA in |
| One® Surgical System (formerly ROBODOC®) [330] | Inc., Fremont, CA, USA | arthroplasty, total knee arthroplasty, and cup placement procedures | instrument that is working in 3D space; preoperative planning system; 3D image display | 2015<br>CE Mark in 2015 |
| Senhance® Surgical System | TransEnterix Inc., Morrisville, NC, USA | Laparoscopic gynecological surgery, colorectal surgery, cholecystectomy, and inguinal hernia repair | Digital laparoscopic platform, tele-manipulation; 3 individual robotic arms: 2 laparoscopic tools (7 DOFs each) with haptic feedback and 1 3D-HD camera | FDA in 2017<br>CE Mark in 2018 |
| Sensei® X Robotic Catheter System [331] | Hansen Medical Inc., Mountain View, CA, USA; was acquired by Auris Surgical Robotics in 2016 | Cardiac therapeutics including cardiac mapping, ablation, and endovascular aneurysm repair | Master-slave tele-manipulation; 3D motion of catheter tip; force sensing, visual and haptic feedback | FDA in 2007 |
| Monarch™ Platform [332] | Auris Health Inc, Redwood City, | Diagnostic and therapeutic | Two robotic arms (6 DOFs each); | FDA in 2018 |
| | CA, USA | bronchoscopic procedures | flexible bronchoscope with an articulated tip; visualization system; one working channel | |
| CyberKnife® System [34] | Accuray Inc., Sunnyvale, CA, USA | Radiosurgical treatment of tumors anywhere in the body | Radiation source is holding by a 6-DOFs robotic arm; real-time tumor tracking system | FDA in 1999<br>CE Mark in 2002 |
| Invendoscopy E210 System [333] | invendo medical GmbH, Kissing, Germany;<br>was acquired by Ambu A/S (Ballerup, Denmark) in 2017 | Diagnostic and therapeutic colonoscopy | A reusable handheld controller; a single-use colonoscope: a 3.2 mm working channel and a flexible tip | FDA in 2018 |
| NeoGuide Colonoscope [334] | NeoGuide Endoscopy System Inc., Los Gatos, CA, USA; | Diagnostic and therapeutic access to the lower gastrointestinal tract | Robotic colonoscope with 16 controlled segments; real- | FDA in 2006 |
| | was acquired by Intuitive Surgical in 2009 | for endoscopy and interventions | time 3D mapping by position sensors | |
| FreeHand v1.2 System [335] | FreeHand 2010 Ltd., Surrey, UK | Urology, gynecology, and general surgery | Manipulating a laparoscope in 3D (pan, tilt, and zoom) by head movement | FDA in 2009 |
| **Surgical robotic systems are under developing** | | | | |
| SPORT Surgical System [336-337] | Titan Medical Inc., Toronto, ON, Canada | Laparoendoscopic single site (LESS) surgery | Master-slave tele-manipulation; 2 multi-articulated controllers; single-arm mobile cart with 2 multi-articulated instruments and 1 3D-HD flexible camera | No |
| MASTER System [338] | EndoMaster Pte. Ltd., Singapore | Natural orifice transluminal endoscopic surgery (NOTES) | Master-slave tele-manipulation; 2 handle interfaces; 2 slave robotic arms (up to 7 DOFs); | No |
| | | | visualization system | |
| SurgiBot [339] | TransEnterix Inc., Morrisville, NC, USA | LESS surgery | Two laparoscopic handles manipulating 2 flexible instruments; 3D-HD visualization; exchangeable instruments | Rejected by FDA in 2016 |
| i-Snake [340] | Imperial College London, UK | Multi-vessel coronary bypass surgery and general diagnosis | Fully articulated joints; “follow the leader” algorithm; multiple sensing at the tip | No |
| Versius® Surgical Robotic System [341-342] | CMR Surgical Ltd., Cambridge, UK | Laparoscopic surgery including gynecology, upper gastrointestinal surgery, colorectal and urology | Master-slave tele-manipulation; 2 joystick controllers with haptic feedback; up to 5 modular fully wristed robotic arms; 3D-HD vision system | CE Mark in 2019 |
| MiroSurge [343- | DLR, Germany | Minimally invasive | Master-slave tele- | No |
| 344] | | robotic telesurgery | manipulation; 2 haptic input devices; 3 individual robotic arms (7 DOFs each): 2 laparoscopic instruments and 1 3D camera; visual and force feedback | |
| Revo-I [345] | meerecompany, Gyeonggi-do, Korea | General laparoscopic surgery including cholecystectomy and prostatectomy | Master-slave tele-manipulation; 2 handle interfaces; 4 robotic arms: 3 instruments (7 DOFs each) and 1 3D-HD camera | Korean FDA in 2017 |
| Surgenius [340] | Surgica Robotica S.r.l, Trieste, Italy | General laparoscopic surgery | Master-slave tele-manipulation; master controllers with haptic feedback; individual 6-DOFs robotic | No |
| | | | arms embedded 6-DOFs tip-force sensors | |